\documentclass{article}
\usepackage{iclr2027_conference,times}

\usepackage{amsmath,amsfonts,bm}

\def\eqref#1{equation~\ref{#1}}

\def\1{\bm{1}}

\DeclareMathAlphabet{\mathsfit}{\encodingdefault}{\sfdefault}{m}{sl}
\SetMathAlphabet{\mathsfit}{bold}{\encodingdefault}{\sfdefault}{bx}{n}

\usepackage[table]{xcolor}
\usepackage{url,booktabs,graphicx,amssymb,tabularx,pifont}
\usepackage{tcolorbox}
\tcbuselibrary{skins}
\usepackage{tikz,pgfplots}
\usetikzlibrary{arrows.meta,positioning,calc,fit,patterns}
\usepgfplotslibrary{groupplots}
\pgfplotsset{compat=1.18}
\usepackage{placeins}
\usepackage{needspace,float}
\usepackage{hyperref}
\usepackage{amssymb}
\definecolor{ctxcolor}{HTML}{226C9A}
\definecolor{memcolor}{HTML}{B66B2A}
\definecolor{cYes}{HTML}{E7F0F6}
\definecolor{cYesT}{HTML}{226C9A}
\definecolor{cNo}{HTML}{FAEBDD}
\definecolor{cNoT}{HTML}{9C571E}
\definecolor{cHead}{HTML}{EDF0F4}
\definecolor{cNa}{HTML}{F3F3F3}
\definecolor{cOa}{HTML}{B66B2A}
\definecolor{cQa}{HTML}{226C9A}
\definecolor{cOb}{HTML}{8B4691}
\definecolor{cQb}{HTML}{278374}
\hypersetup{colorlinks=true,linkcolor=ctxcolor,citecolor=ctxcolor,urlcolor=ctxcolor}
\newcommand{\ctx}{\texttt{context}}
\newcommand{\kno}{\texttt{knowledge}}

\newcommand{\from}{\ensuremath{\leftarrow}}
\newcommand{\dt}{\ensuremath{\,\cdot\,}}
\newcommand{\olmoS}{OLMo-3-7B}
\newcommand{\olmoL}{OLMo-3.1-32B}
\newcommand{\qwenS}{Qwen3-8B}
\newcommand{\qwenL}{Qwen3-32B}
\newcommand{\xp}[1]{\textup{\textsc{#1}}}
\newcommand{\xDrop}{\xp{Document drop}}
\newcommand{\xDswap}{\xp{Document swap}}
\newcommand{\xQswap}{\xp{Question swap}}

\newcommand{\xSsteer}{\xp{Source steering}}
\newcommand{\xIswap}{\xp{Identical-text swap}}
\newcommand{\cYesCell}[1]{\cellcolor{cYes}\textcolor{cYesT}{\ding{51}}~#1}
\newcommand{\cNoCell}[1]{\cellcolor{cNo}\textcolor{cNoT}{\ding{55}}~#1}

\newtcolorbox{examplebox}[1]{enhanced,colback=cHead!45,colframe=ctxcolor!50,
 boxrule=.45pt,arc=1.5pt,left=6pt,right=6pt,top=4pt,bottom=4pt,
 fontupper=\small,title={#1},fonttitle=\small\bfseries,colbacktitle=cHead,coltitle=black}
\pgfplotsset{paperaxis/.style={font=\footnotesize,axis line style={black!40},
 tick style={black!40},grid style={black!10},axis x line*=bottom,axis y line*=left,
 tick label style={font=\scriptsize},label style={font=\footnotesize},
 title style={font=\small\bfseries},clip=false}}

  \iclrfinalcopy
\title{Context or Knowledge:\\How LLMs Attribute Their Own Answers}
\author{Kirill Afendulev \quad Alexey Dontsov \quad Alla Chepurova\\[4pt]
  Elena Tutubalina \quad Anton Korznikov}
\hypersetup{
  pdftitle={Context or Knowledge: How LLMs Attribute Their Own Answers},
  pdfauthor={Kirill Afendulev, Alexey Dontsov, Alla Chepurova, Elena Tutubalina, Anton Korznikov}
}

\makeatletter
\renewcommand{\@maketitle}{%
  \vbox{\hsize\textwidth
    \centering
    {\LARGE\bfseries\@title\par}
    \vskip 14pt
    {\normalsize\@author\par}
    \vskip 18pt
  }%
}
\makeatother

\begin{document}
\raggedbottom
\maketitle

\begin{abstract}
Language models can answer a question from a supplied document or from their parametric knowledge. When later asked where the answer came from, what determines their source report? We study this by intervening on the model's key--value (KV) cache after the answer is generated and before the source question is asked. On a controlled factual task with four models (7B--32B), changing or removing the document's cache never changes the report. In contrast, transplanting the question-and-answer (Q+A) cache between runs with identical question and answer tokens transfers it: every eligible report follows the donor, while the recipient's document is unchanged. On the two smaller models, steering Q+A values along directions fitted before any source question reverses 98.9--100\% of reports across four document conditions at a fixed strength, while norm-matched random directions flip at most 1.1\%. The components of keys and values along fitted directions carry the transplant effect: copying them alone nearly reproduces the report change; holding them fixed nearly prevents it. Applied during answer generation, the same value directions also change which source supplies the answer, though more weakly and depending on the prompt. Source reporting in this setting thus depends causally on cached state produced while answering. We do not establish general reporting accuracy or a complete interpretation of this state.

\end{abstract}

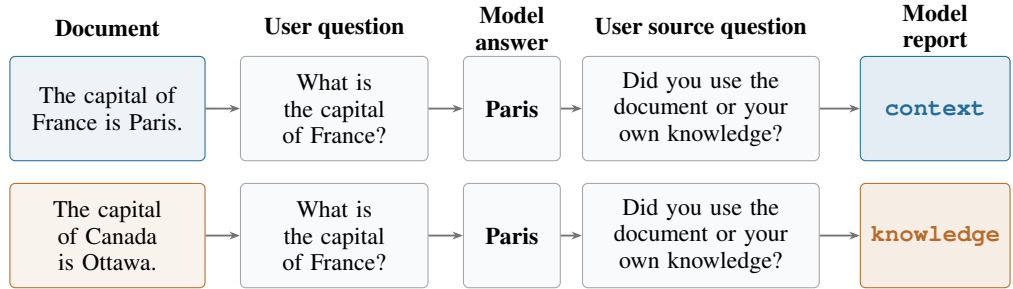
\begin{figure}[H]
\centering
\begin{tikzpicture}[
 block/.style={draw=black!35,fill=cHead!35,rounded corners=2pt,
   minimum height=1.38cm,inner sep=4pt,align=center,font=\footnotesize},
 heading/.style={align=center,font=\footnotesize\bfseries},
 flow/.style={-{Stealth[length=4pt]},line width=.7pt,black!55}]

\node[heading] at (0,1.06) {Document};
\node[heading] at (3,1.06) {User question};
\node[heading] at (5.35,1.06) {Model\\answer};
\node[heading] at (7.85,1.06) {User source question};
\node[heading] at (10.95,1.06) {Model\\report};

\foreach \y/\name/\col/\fact/\report in {
  0/a/ctxcolor/{The capital of France is Paris.}/{\ctx{}},
  -1.68/b/memcolor/{The capital of Canada is Ottawa.}/{\kno{}}} {
  \node[block,text width=2.3cm,draw=\col,fill=\col!8] (\name-doc) at (0,\y) {\fact};
  \node[block,text width=2.2cm] (\name-q) at (3,\y) {What is the capital of France?};
  \node[block,text width=1cm,font=\small\bfseries] (\name-a) at (5.35,\y) {Paris};
  \node[block,text width=2.85cm] (\name-source) at (7.85,\y)
    {Did you use the document or your own knowledge?};
  \node[block,text width=1.7cm,draw=\col,fill=\col!12,text=\col,font=\small\bfseries]
    (\name-report) at (10.95,\y) {\report};
  \draw[flow] (\name-doc.east) -- (\name-q.west);
  \draw[flow] (\name-q.east) -- (\name-a.west);
  \draw[flow] (\name-a.east) -- (\name-source.west);
  \draw[flow] (\name-source.east) -- (\name-report.west);
}
\end{tikzpicture}
\caption{\textbf{Same answer, different source reports.} Each row is a separate dialogue. After answering, the model reports \ctx{} for the document or \kno{} for its own knowledge. We ask how this report is produced. Schematic examples; source-question wording shortened.}
\label{fig:dialogue}
\end{figure}

\section{Introduction}
\label{sec:intro}

When a language model answers a question with a document in context, a user may want to know where the answer came from. Did the model use the document, or what it already knew? This matters when the document updates or contradicts the model's knowledge. We can ask the model directly (Figure~\ref{fig:dialogue}). Asked for France's capital, it answers ``Paris'' with either a supporting or an irrelevant document, but reports \ctx{} in the first case and \kno{} in the second. We call this reply the \emph{source report}. Our research question is how the model produces it.

This question is distinct from asking whether the document supports the answer. A source report is a claim about \emph{how the answer was produced}. In retrieval-augmented systems, both a retrieved document and parametric knowledge may support the same answer. Prior work on attribution evaluates whether generated content is supported by the available context~\citep{rashkin2023measuring}; work on explanation faithfulness likewise shows that a model's verbal account of its computation need not directly expose the computation itself~\citep{turpin2023language,lanham2023measuring}. Appendix~\ref{app:related} reviews related work in more detail. We instead ask what internal state the model uses when it produces a source report about an answer that has already been generated.

Our starting point is the transformer's key--value (KV) cache. During generation, previously processed tokens leave keys and values that later tokens can attend to. These entries contain more than the token identities themselves: the same question tokens can acquire different cached states depending on the document that was processed before them. This gives us a way to separate \emph{what the model saw} from \emph{the state produced by processing it}. We let the model answer normally, intervene on selected cache entries, and only then ask it about the source. This setup lets us distinguish three possible mechanisms for the source report:

\textbf{H1: Re-read the document.}
The model may inspect the document again when asked about the source. Under this hypothesis, changing the document after answering should change the report. 

\textbf{H2: Judge the answer using parametric knowledge.} The model may instead assess whether its answer is compatible with knowledge stored in its parameters. Under this hypothesis, the report can be determined from the answer's compatibility with parametric knowledge rather than from a record of how the answer was produced. 

\textbf{H3: Read a cached trace of the earlier computation.} The model may use source-related information left in the cache while answering. If so, transplanting the Q+A cache between runs should transfer the report, even when the recipient's question, answer, and document stay unchanged.

\begin{table}[!htbp]
\caption{\textbf{Hypotheses for the source report and evidence from our interventions.} 
Blue \textcolor{cYesT}{\ding{51}} marks evidence supporting the hypothesis; orange \textcolor{cNoT}{\ding{55}} marks evidence against it. H1 and H2 are tested as sufficient explanations of the report. H3 represents the hypothesis that the report depends on source-related state produced while answering.}
\label{tab:hyp}
\centering\footnotesize\setlength{\tabcolsep}{4pt}\renewcommand{\arraystretch}{1.1}
\begin{tabularx}{\linewidth}{@{}>{\raggedright\arraybackslash}p{0.8in}>{\raggedright\arraybackslash}p{1.4in}>{\raggedright\arraybackslash}X@{}}
\toprule
\rowcolor{cHead}
\textbf{Hypothesis} & \textbf{Proposed mechanism} & \textbf{Evidence} \\
\midrule

\textbf{H1: Re-read the document} &
The model re-inspects the document to produce the source report. &
\textbf{\cNoCell{} Not supported as a sufficient explanation.}
Changing or removing the document after answering does not change the report
(\S\ref{sec:b1}). \\

\addlinespace[3pt]
\textbf{H2: Judge the answer against knowledge} &
The model determines the report by evaluating the answer against its parametric knowledge. &
\textbf{\cNoCell{} Not supported as a sufficient explanation.}
Transplanting the Q+A cache changes the report while keeping the answer tokens and recipient document unchanged
(\S\ref{sec:b2}). \\

\addlinespace[3pt]
\textbf{H3: Read a cached trace} &
The report depends on source-related state produced while answering. &
\textbf{\cYesCell{} Supported by our interventions.}
Q+A cache transplants transfer the report, and steering Q+A values changes it
(\S\ref{sec:b2}, \S\ref{sec:b3}). Clamping fitted K+V projections nearly blocks transfer; transferring those projections alone nearly reproduces it
(\S\ref{sec:clamp}). \\

\bottomrule
\end{tabularx}
\end{table}

We test these mechanisms in sequence (Table~\ref{tab:hyp}). First, we change or remove the document's cache after the answer has been generated while preserving the Q+A cache. The report remains unchanged, even though the replacement document is still readable. This argues against determining the report from the document currently available at report time. Second, we transplant the Q+A cache between runs with identical question and answer tokens. The report follows the transplanted cache, while the recipient's document remains fixed. This establishes a causal dependence on the state produced during the earlier computation. Third, we fit linear directions in Q+A values and show that steering these values can change the source report across facts. Fitted K+V components also carry nearly the full transplant effect: transferring them nearly reproduces the report change, while holding them fixed nearly prevents it (Section~\ref{sec:clamp}). Finally, we apply the same fitted value directions during answer generation, without refitting. They change which source supplies the answer, linking source-related cached state to source selection even under opposing instructions.

Importantly, H1 and H2 are tested as \emph{sufficient explanations} of the report, not as claims that document processing and parametric knowledge cannot contribute to the cached state. A successful Q+A transplant does not imply that the trace represents a single discrete ``source label''; it shows that the state produced during the earlier computation contains information that causally affects the later report. Hybrid mechanisms may therefore contribute to the computation as well.

Our main contribution is therefore a mechanistic explanation of source reporting in the controlled setting: the source report depends on source-related computational state inherited from the question-answering process, rather than simply on the document that remains available when the report is requested. We call this state a \emph{source trace}; its role in reporting leaves open whether it encodes evidence availability, actual source use, or both. The experiments do not establish general reliability of source reports, unrestricted introspection, or a complete semantic interpretation of this state.

\section{Setup: hold the words fixed, change their cached state}
\label{sec:setup}
\paragraph{Facts and documents.} We start from 155 candidate facts across eight relations: capital, continent, official language, chemical symbol, author, painter, composer, and inventor. Each fact has seven one-sentence documents: a true fact, two counterfactuals, two irrelevant facts, and two neutral sentences. Document variants have equal token counts within each item under each model's tokenizer. Every later token therefore sits at the same position in each run of that fact, so cache entries can be copied between runs without padding.

\paragraph{Models.} We test \olmoS{} and \olmoL{} \citep{olmo2025olmo3}, and \qwenS{} and \qwenL{} \citep{yang2025qwen3}, with Qwen's thinking disabled. Every dialogue is shorter than 300 tokens and fits each model's attention window. Steering experiments use the two smaller models.

\begin{examplebox}{One question, different routes to an answer (schematic)}
\small
\setlength{\tabcolsep}{4pt}\renewcommand{\arraystretch}{1.05}
\begin{tabularx}{\linewidth}{@{}l>{\raggedright\arraybackslash}Xll@{}}
\multicolumn{4}{@{}l}{\textbf{Question (every run):} What is the capital of France?}\\[3pt]
\textbf{Type} & \textbf{Document} & \textbf{Answer} & \textbf{Intact report}\\
\midrule
True fact      & ``The capital of France is Paris.''  & Paris & \ctx{}\\
Counterfactual & ``The capital of France is Tokyo.''  & Tokyo & \ctx{}\\
Irrelevant     & ``The capital of Canada is Ottawa.'' & Paris & \kno{}\\
Neutral        & ``Snow melted slowly on the hills.'' & Paris & \kno{}\\
\end{tabularx}\par\smallskip
{\footnotesize\textbf{Source question:} ``What is the source of the answer you just gave: the provided Document, or your own knowledge? Answer `context' for the Document or `knowledge' for your own knowledge. One word only.''\par}
\end{examplebox}

\paragraph{Dialogue and source report.} The model sees a document and a question, and generates its answer greedily. In Sections~\ref{sec:b1}--\ref{sec:clamp}, the LLM is instructed to use a relevant document even when it conflicts with its knowledge, and otherwise to answer from knowledge. The fixed-document follow-up (Appendix~\ref{app:fixedcontext}) and the experiment in Section~\ref{sec:b4} vary this instruction explicitly. After the answer, we ask the source question. We call the model's reply the \emph{source report}: \ctx{} for the document, \kno{} for its own knowledge. The report is the label with the higher summed token log-probability. This agrees with the model's freely generated one-word reply in 99.6--100\% of cases.

\paragraph{Eligible runs.} The \emph{intact report} is the report of an unedited run. We keep only \emph{eligible} runs, whose answer and intact report match those expected for their document type (above), using rules fixed before the experiments. Eligibility is 92--100\% by model and condition. The exception is irrelevant-fact runs on \olmoS{} (81--82\%, mainly refusals). This filter ensures that every edited run starts from a known answer and report, so that a change can be attributed to the edit. Reported success rates are therefore conditional on eligibility.

\paragraph{Intervention timing.} For every token and layer the KV cache stores the key and value vectors that later tokens attend to. We divide the dialogue into spans. The \emph{document} span is the document tokens. The \emph{question span} (Q) contains the question and the template tokens between the document and the answer. The \emph{answer span} (A) contains the generated answer tokens. We edit the cache \emph{after} the answer is generated and \emph{before} the source question is processed. We use two types of edits:

(1) A \emph{drop} removes the document's cache entries. Later tokens keep their original position indices.

(2) A \emph{swap} (or \emph{transplant}) replaces the keys and values of a span with those from another run of the same fact that saw a different document. We write ``recipient~$\leftarrow$~donor'': the recipient run receives the donor's cache. Swapping the answer span requires identical answer tokens in both runs. Our main transplant replaces Q and A together (Q+A).
Entries we do not edit are kept, not recomputed. A token's text thus stays the same while its cached state can come from a different processing history.

\paragraph{Outcomes and rates.} A \emph{flip} is a change of the source report to the other label. A \emph{source switch} (Section~\ref{sec:b4}) is a change of the answer to the other source's value, e.g., from ``Tokyo'' to ``Paris''. Rates in Sections~\ref{sec:b1}--\ref{sec:b2} are over eligible facts. Steering directions are fitted on some relations and evaluated on others. Report-steering rates pool two retained splits (Appendix~\ref{app:strength}); steering rates are over \emph{item--split pairs}, not independent facts. The examples are schematic illustrations, not quoted transcripts. Detailed rates appear in Appendix~\ref{app:documents} onward.

\Needspace{8\baselineskip}
\section{Changing the document leaves the report in place}
\label{sec:b1}

\begin{figure}[H]
\centering
\begin{tikzpicture}[
 doc/.style={draw=ctxcolor,fill=ctxcolor!8,rounded corners=2pt,
   text width=3.20cm,minimum height=1.04cm,inner sep=5pt,align=center,font=\small},
 qa/.style={draw=ctxcolor,line width=.8pt,fill=ctxcolor!8,rounded corners=3pt,
   minimum width=3.62cm,minimum height=1.85cm},
 turn/.style={text width=3.05cm,inner sep=3pt,align=center,font=\small},
 source/.style={draw=black!35,fill=cHead!40,rounded corners=2pt,
   text width=3.20cm,minimum height=.78cm,inner sep=5pt,align=center,font=\footnotesize},
 report/.style={draw=ctxcolor,fill=ctxcolor!12,text=ctxcolor,
   text width=3.20cm,minimum height=.45cm,inner sep=4pt,align=center,font=\small\bfseries},
 flow/.style={-{Stealth[length=4pt]},line width=.7pt,black!55},
 copy/.style={-{Stealth[length=5pt]},line width=1pt,ctxcolor}]

\foreach \x/\name/\heading in {0/a/{1. Original dialogue},4.55/b/{2. Drop document},9.10/c/{3. Replace document}} {
  \node[font=\small\bfseries,align=center] at (\x,.48) {\heading};
  \node[qa] (\name-qa) at (\x,-2.27) {};
  \node[turn] (\name-q) at (\x,-1.88)
    {\textbf{User:} What is the\\capital of France?};
  \draw[ctxcolor!30] (\x-1.57,-2.33) -- (\x+1.57,-2.33);
  \node[turn] (\name-a) at (\x,-2.68) {\textbf{Model:} Tokyo};
  \node[source] (\name-source) at (\x,-4.05)
    {\textbf{User:} Answer source?\\\ctx{} / \kno{}};
  \node[report] (\name-report) at (\x,-5.06) {\ctx{}};
  \draw[flow,ctxcolor] (\name-qa.south) -- (\name-source.north);
  \draw[flow] (\name-source.south) -- (\name-report.north);
}

\node[doc] (a-doc) at (0,-.38)
  {\textbf{Document}\\The capital of France is Tokyo.};
\node[doc,draw=black!40,dashed,fill=black!3,text=black!65] (b-doc) at (4.55,-.38)
  {\textbf{No document KV}\\Entries removed};
\node[doc,draw=memcolor,fill=memcolor!8] (c-doc) at (9.10,-.38)
  {\textbf{Replacement document}\\The capital of Canada is Ottawa.};

\draw[flow,ctxcolor] (a-doc.south) -- (a-qa.north);
\draw[copy] (a-qa.east) -- (b-qa.west)
  node[midway,above=2pt,font=\scriptsize,align=center] {copy\\KV};
\draw[copy] (b-qa.east) -- (c-qa.west)
  node[midway,above=2pt,font=\scriptsize,align=center] {same\\KV};
\draw[flow,memcolor] (c-doc.east) -- ++(.28,0) |- (c-source.east);

\node[font=\scriptsize,text=ctxcolor,align=center] at (4.55,-5.64)
  {Blue Q+A state is preserved; the question and answer are not recomputed.};
\end{tikzpicture}
\caption{\textbf{The document changes; the source report does not.} The LLM answers ``Tokyo'' with a counterfactual document. Each edited branch keeps the original question-and-answer cache, and either removes the document entries or replaces them with an irrelevant fact, which remains readable. All three branches then report \ctx{} to the source question, and the answer shown was generated before either edit. Schematic example; Table~\ref{tab:document_main} gives results for all four models.}
\label{fig:document_edits}
\end{figure}
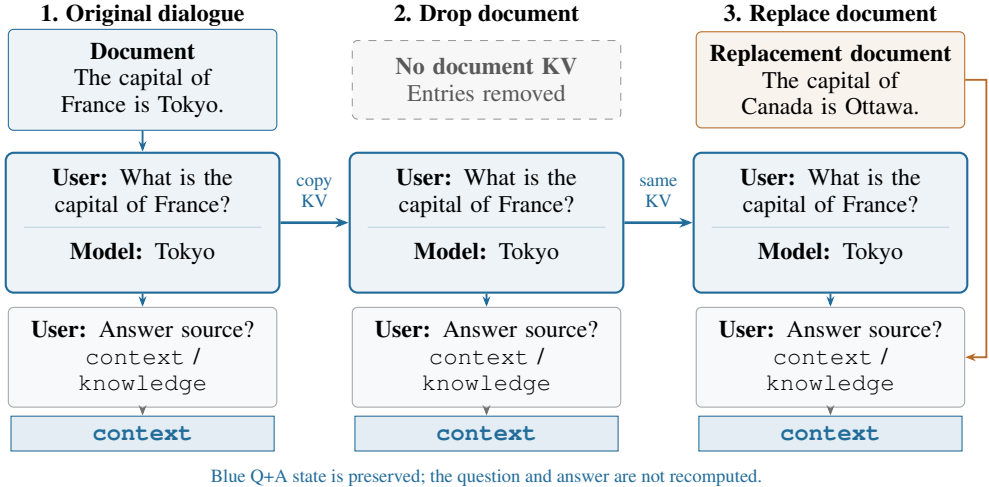

Under H1, the report may compare the answer with the document currently in context. If so, changing the document after the model has answered should change the report. The model reads ``The capital of France is Tokyo'' and answers ``Tokyo''. We keep its question-and-answer cache and either remove the document's entries or replace them with an irrelevant fact. Only then do we ask the source question (Figure~\ref{fig:document_edits}).

\begin{table}[H]
\caption{\textbf{Zero source-report flips after document edits.} Cells are report-flip rates (\%), with Q+A state retained. The first two edits try to remove a \ctx{} report; the third tries to create one.}
\label{tab:document_main}
\centering\footnotesize\setlength{\tabcolsep}{5pt}
\begin{tabular}{@{}lcccc@{}}
\toprule
Document edit & \olmoS{} & \qwenS{} & \olmoL{} & \qwenL{}\\
\midrule
Drop counterfactual & 0\% & 0\% & 0\% & 0\%\\
Counterfactual $\to$ irrelevant & 0\% & 0\% & 0\% & 0\%\\
Irrelevant $\to$ true fact & 0\% & 0\% & 0\% & 0\%\\
\bottomrule
\end{tabular}
\end{table}

In this setting, neither edit flips a report on any model. The edit is available to the model: in every eligible run with a replaced document, the model can quote the new document. The reverse test leads to the same conclusion. After the model answers ``Paris'' from memory, inserting a true fact stating ``Paris'' into the document never turns \kno{} into \ctx{}. The report therefore does not track whether the current document supports the answer. DynamicQA Wikipedia paragraphs show the same pattern, with rare exceptions (Appendix~\ref{app:dynamicqa}). Table~\ref{tab:b1} gives the full templated-task results.

\Needspace{8\baselineskip}
\section{The report follows the question-and-answer cache}
\label{sec:b2}
The previous experiment showed that the report does not change when we replace the document after the answer has been generated. This suggests that the information used for reporting may be retained in the model's internal state from the earlier question-answering computation. We test this by transplanting the KV cache of the question and answer (\emph{Q+A cache}) between runs.

\begin{figure}[!htbp]
\centering
\begin{tikzpicture}[
 doc/.style={draw=ctxcolor,fill=ctxcolor!8,rounded corners=2pt,
   text width=3.20cm,minimum height=1.04cm,inner sep=5pt,align=center,font=\small},
 qa/.style={draw=ctxcolor,line width=.8pt,fill=ctxcolor!8,rounded corners=3pt,
   minimum width=3.62cm,minimum height=1.85cm},
 turn/.style={text width=3.05cm,inner sep=3pt,align=center,font=\small},
 source/.style={draw=black!35,fill=cHead!40,rounded corners=2pt,
   text width=3.20cm,minimum height=.78cm,inner sep=5pt,align=center,font=\footnotesize},
 report/.style={draw=ctxcolor,fill=ctxcolor!12,text=ctxcolor,
   text width=3.20cm,minimum height=.45cm,inner sep=4pt,align=center,font=\small\bfseries},
 flow/.style={-{Stealth[length=4pt]},line width=.7pt,black!55},
 copy/.style={-{Stealth[length=5pt]},line width=1pt,ctxcolor}]

\foreach \x/\name/\heading/\col in {0/a/{1. Original recipient}/memcolor,4.55/b/{2. Donor}/ctxcolor,9.10/c/{3. Edited recipient}/ctxcolor} {
  \node[font=\small\bfseries,align=center] at (\x,.48) {\heading};
  \node[qa,draw=\col,fill=\col!8] (\name-qa) at (\x,-2.27) {};
  \node[turn] (\name-q) at (\x,-1.88)
    {\textbf{User:} What is the\\capital of France?};
  \draw[\col!30] (\x-1.57,-2.33) -- (\x+1.57,-2.33);
  \node[turn] (\name-a) at (\x,-2.68) {\textbf{Model:} Paris};
  \node[source] (\name-source) at (\x,-4.05)
    {\textbf{User:} Answer source?\\\ctx{} / \kno{}};
  \draw[flow,\col] (\name-qa.south) -- (\name-source.north);
}

\node[doc,draw=memcolor,fill=memcolor!8] (a-doc) at (0,-.38)
  {\textbf{Neutral document}\\Snow melted slowly on the hills.};
\node[doc] (b-doc) at (4.55,-.38)
  {\textbf{True-fact document}\\The capital of France is Paris.};
\node[doc,draw=memcolor,fill=memcolor!8] (c-doc) at (9.10,-.38)
  {\textbf{Document unchanged}\\Snow melted slowly on the hills.};

\node[report,draw=memcolor,fill=memcolor!12,text=memcolor] (a-report) at (0,-5.06) {\kno{}};
\node[report] (b-report) at (4.55,-5.06) {\ctx{}};
\node[report] (c-report) at (9.10,-5.06) {\ctx{}};
\foreach \name in {a,b,c} {
  \draw[flow] (\name-source.south) -- (\name-report.north);
}

\draw[flow,memcolor] (a-doc.south) -- (a-qa.north);
\draw[flow,ctxcolor] (b-doc.south) -- (b-qa.north);
\draw[copy] (b-qa.east) -- (c-qa.west)
  node[midway,above=2pt,font=\scriptsize,align=center] {copy\\KV};
\draw[flow,memcolor] (c-doc.east) -- ++(.28,0) |- (c-source.east);

\node[font=\scriptsize,align=center] at (4.55,-5.64)
  {Same recipient words; \textcolor{memcolor}{original Q+A state} replaced by \textcolor{ctxcolor}{donor Q+A state}.};
\end{tikzpicture}
\caption{\textbf{Source report follows the transplanted cache.} Both runs answer ``Paris'', but report different sources. We copy the donor's Q+A cache into the recipient after answering, retaining the recipient's document. Asked anew, the source question now elicits \ctx{}. Colours identify the origin of cached state; the original question and its answer are not regenerated.}
\label{fig:identswap}
\end{figure}
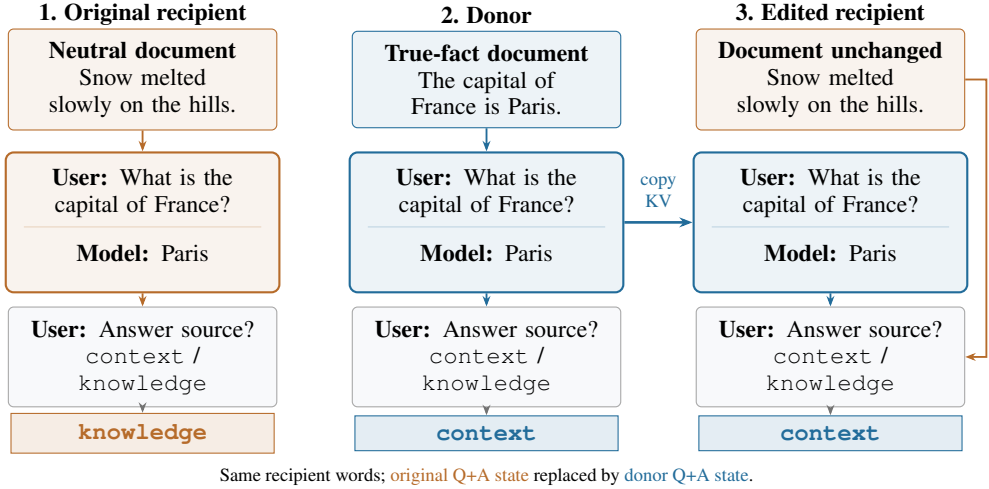

We pair runs with identical question and answer tokens but different documents. For example, both runs may answer ``Paris'', one with a supporting document and a \ctx{} report, the other with a neutral document and a \kno{} report (Figure~\ref{fig:identswap}). We copy the Q+A cache from the \emph{donor} run into the \emph{recipient}; the recipient's document and all other cache entries remain unchanged.

After the transplant, the report consistently follows the donor's cache across all four tested models (Table~\ref{tab:transplant_main}). For example, replacing the Q+A cache of a neutral or irrelevant-document run with the cache from a supporting-document run changes the recipient's report to \ctx{}, even though the recipient's original document is not relevant. Conversely, transplanting the Q+A cache from a neutral or irrelevant-document run causes a recipient that still contains supporting evidence to report \kno{}. Control transplants, which preserve the source condition, produce no report flips. On DynamicQA Wikipedia paragraphs, joint Q+A transplants similarly produce 97.2--100\% donor-following (Appendix~\ref{app:dynamicqa}).

\begin{table}[!htbp]
\caption{\textbf{Every eligible report follows the Q+A donor.} Cells are report-flip rates (\%) among eligible facts with identical answer tokens. The document stays unchanged; only Q+A KV entries are replaced.}
\label{tab:transplant_main}
\centering\footnotesize\setlength{\tabcolsep}{5pt}
\begin{tabular}{@{}lcccc@{}}
\toprule
Recipient $\leftarrow$ Q+A donor & \olmoS{} & \qwenS{} & \olmoL{} & \qwenL{}\\
\midrule
True fact $\leftarrow$ neutral & 100\% & 100\% & 100\% & 100\%\\
Neutral $\leftarrow$ true fact & 100\% & 100\% & 100\% & 100\%\\
True fact $\leftarrow$ irrelevant & 100\% & 100\% & 100\% & 100\%\\
Irrelevant $\leftarrow$ true fact & 100\% & 100\% & 100\% & 100\%\\
\bottomrule
\end{tabular}
\par\smallskip\raggedright\scriptsize
Re-splicing the recipient's own cache or using a same-condition donor: zero flips on every model.
\end{table}

The intervention changes the report while leaving the recipient's document and answer unchanged. Thus, the report cannot be determined solely by whether the recipient's document supports the answer, because that document is identical before and after the transplant. Likewise, the answer itself is identical, so the report cannot be explained simply by the answer string or by a difference in the model's final answer. Instead, the intervention changes only the Q+A cache, indicating that source-related information carried in this state contributes to the report.

\paragraph{Why transplant the question and answer together?} The source-related information may be distributed across the computation of both the question and the answer. We therefore transplant their KV entries jointly, which produces a consistent effect across conditions. Results for separate question-only and answer-only transplants are reported in Appendices~\ref{app:transplants} and~\ref{app:where}.

The transplant experiment therefore provides causal evidence that the later source report depends on information encoded in the Q+A cache. We next ask whether this source-related component can be directly manipulated by a shared intervention in the cache.

\Needspace{12\baselineskip}
\section{Steering question-and-answer values changes the report}
\label{sec:b3}

We next ask whether the source-related state identified by the transplant experiment can be directly manipulated. We construct directions in Q+A value space from the difference between runs with supporting and irrelevant documents.

For each model and relation split, we fit directions on one set of relations and evaluate them on the remaining relations. Fitting examples pair a \emph{supporting-fact} document that supports the answer with an \emph{irrelevant-fact} document that does not; answer-span fitting requires identical answer tokens. At each layer, we average the value vectors over the positions of the question or answer span and compute the mean difference between the two conditions:

\begin{equation}
 d^{s}_{\ell}=\mathop{\mathrm{mean}}_{i\in\mathcal F_s}
 \left(\overline V^{\,\mathrm{true},s}_{\ell,i}-\overline V^{\,\mathrm{irrelevant},s}_{\ell,i}\right),
 \quad V'_{\ell,t}=V_{\ell,t}\pm\alpha d^{s(t)}_{\ell},
 \quad s\in\{Q,A\}.
 \label{eq:direction}
\end{equation}

Here, $s$ specifies whether the intervention is applied to the question or answer span, and $\mathcal F_s$ is the corresponding fitting set. Values carry the information retrieved by attention; the key/value ablations motivating this choice are in Appendix~\ref{app:kv}. Fitting uses states recorded before any source question. We then intervene on the Q+A values after the model has already produced its answer, leaving the keys and all other cache entries unchanged. Adding the direction moves the Q+A state toward the supporting-document condition; subtracting it moves the state in the opposite direction. At $\alpha=1$, the intervention has the magnitude of the mean difference between the two fitting conditions.

As a control, we apply random directions with the same layerwise norms and the same intervention locations and strengths. Each random condition is averaged over three draws.

\begin{examplebox}{Changing the report without changing the answer (schematic)}
\textbf{Completed dialogue:} the model has answered ``Tokyo'' to a question.\\ \textbf{Intervention:} after the answer is generated, subtract the learned source direction from the Q+A values.\\ \textbf{Result:} the source report changes from \ctx{} to \kno{}, while the previously generated answer remains ``Tokyo''.
\end{examplebox}

\begin{figure}[!htbp]
\centering
\begin{tikzpicture}
\begin{groupplot}[group style={group size=2 by 1,horizontal sep=0.75cm},
 paperaxis,scale only axis,width=5.05cm,height=2.4cm,
 ymin=-3,ymax=107,ytick={0,50,100},ymajorgrids,
 xmin=.8,xmax=3.2,xtick={1,2,3},xlabel={Strength $\alpha$}]
\nextgroupplot[title={Create a context report},ylabel={Reports flipped (\%)}]
\addplot[cOa,thick,mark=*,mark size=2.5pt] coordinates {(1.000000,42.622951) (2.000000,100.000000) (3.000000,100.000000)};
\addplot[cOa,dotted,thick] coordinates {(1.000000,0.000000) (2.000000,0.000000) (3.000000,1.147541)};
\addplot[cQa,thick,mark=square,mark size=2.5pt] coordinates {(1.000000,4.938272) (2.000000,100.000000) (3.000000,100.000000)};
\addplot[cQa,dotted,thick] coordinates {(1.000000,0.000000) (2.000000,0.000000) (3.000000,0.000000)};
\nextgroupplot[title={Remove a context report}]
\addplot[cOa,thick,mark=*,mark size=2.5pt] coordinates {(1.000000,0.000000) (2.000000,60.465116) (3.000000,100.000000)};
\addplot[cOa,dotted,thick] coordinates {(1.000000,0.000000) (2.000000,0.000000) (3.000000,0.000000)};
\addplot[cQa,thick,mark=square,mark size=2.5pt] coordinates {(1.000000,12.500000) (2.000000,100.000000) (3.000000,100.000000)};
\addplot[cQa,dotted,thick] coordinates {(1.000000,0.000000) (2.000000,0.000000) (3.000000,0.000000)};

\end{groupplot}
\end{tikzpicture}
\par\smallskip\footnotesize
\textcolor{cOa}{$\bullet$ OLMo-3-7B}\quad
\textcolor{cQa}{$\square$ Qwen3-8B}\quad Dotted: matched random
\caption{\textbf{Steering Q+A values changes the source report.} Left: addition makes irrelevant-document runs report \ctx{}; both reach 100\% at $\alpha=2$. Right: subtraction makes counterfactual runs report \kno{}; both reach 100\% by $\alpha=3$. The answer tokens remain fixed. Dotted curves show mean random-control rates over three draws, at most 1.1\% on the left and zero on the right. Rates are over eligible item--split pairs. Other document conditions and span ablations appear in Tables~\ref{tab:b3b}--\ref{tab:b3d}.}
\label{fig:report}
\end{figure}
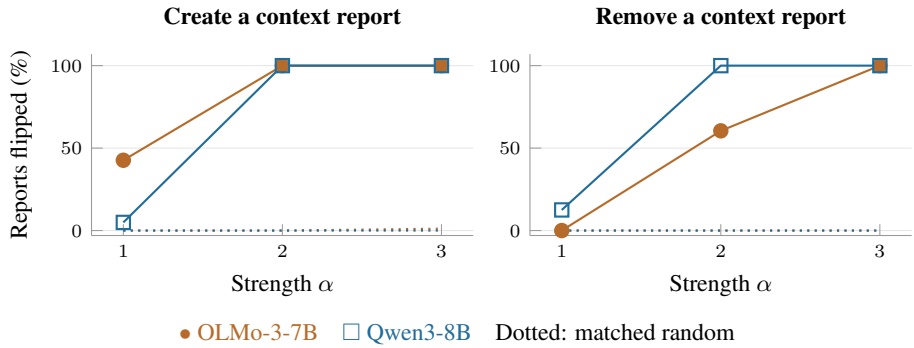

The learned direction reliably changes the subsequent source report in both directions (Figure~\ref{fig:report}). With an irrelevant document, adding the direction produces a \ctx{} report in 100\% of evaluations at $\alpha=2$ on both models. Conversely, with a counterfactual document, subtracting the direction removes the \ctx{} report in 100\% of evaluations by $\alpha=3$ on both models. In all of these interventions, the answer tokens were generated before the edit and remain fixed while the subsequent source report changes.

The effect also generalises to neutral and true-fact documents. Across all four document conditions, $\alpha=3$ flips 100\% of reports except on OLMo true facts (98.9\%). At $\alpha=1$--3, random controls produce no flips on Qwen and at most 1.1\% on OLMo. Additional results, including the comparison of joint and separate-span interventions, are in Appendices~\ref{app:strength}--\ref{app:b3d}.

These results show that the Q+A cache contains a component that can be accessed by a shared linear intervention to change the source report. They do not imply that the entire source-related state is one-dimensional: the learned direction captures a manipulable component of the natural Q+A state.

For irrelevant ($\alpha=2$) and counterfactual documents ($\alpha=3$), every report flips while answer repetition and document quotation both succeed. Both models also retrieve the counterfactual value in every case. These checks use separate continuations of the same edited cache (Appendix~\ref{app:edits}).

\paragraph{Interpretation of the contrast.}
This contrast does not separate evidence availability from source use. A fixed-document contrast also supports report steering (Appendix~\ref{app:fixedcontext}), but neither contrast establishes a pure use signal. The intervention therefore establishes that the state produced during answering is sufficient to change the later source report.

\section{Fitted directions capture the report-transfer effect}
\label{sec:clamp}

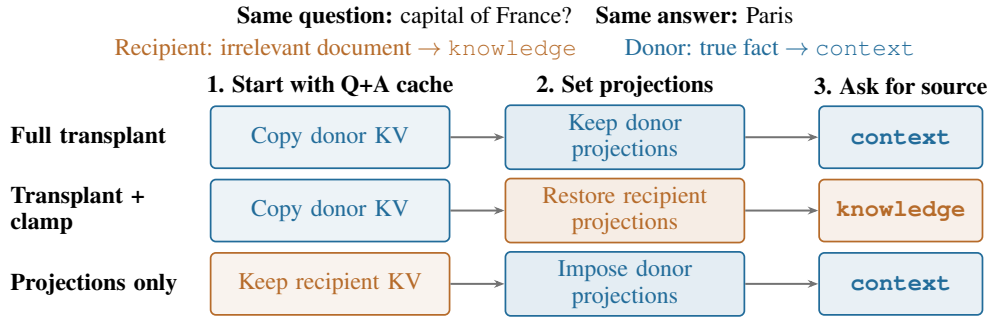
\begin{figure}[!htbp]
\centering
\begin{tikzpicture}[
  font=\footnotesize,
  action/.style={rounded corners=2pt,line width=.65pt,
    text width=2.95cm,minimum height=.82cm,inner sep=3pt,align=center},
  report/.style={rounded corners=2pt,line width=.65pt,
    text width=1.95cm,minimum height=.82cm,inner sep=3pt,
    align=center,font=\footnotesize\bfseries},
  flow/.style={-{Stealth[length=4pt]},black!55,line width=.75pt}]

\node[align=center,font=\footnotesize] at (6.8,.68)
  {\textbf{Same question:} capital of France?\quad \textbf{Same answer:} Paris\\[2pt]
   \textcolor{memcolor}{Recipient: irrelevant document $\to$ \kno{}}\qquad
   \textcolor{ctxcolor}{Donor: true fact $\to$ \ctx{}}};

\node[font=\footnotesize\bfseries] at (4.35,-.02) {1. Start with Q+A cache};
\node[font=\footnotesize\bfseries] at (8.25,-.02) {2. Set projections};
\node[font=\footnotesize\bfseries,align=center] at (11.9,-.02) {3. Ask for source};

\foreach \y/\id/\heading/\base/\operation/\coord/\coordtext/\out/\answer in {
  -.70/a/{Full transplant}/ctxcolor/{Copy donor KV}/ctxcolor/{Keep donor\\projections}/ctxcolor/context,
  -1.67/b/{Transplant + clamp}/ctxcolor/{Copy donor KV}/memcolor/{Restore recipient\\projections}/memcolor/knowledge,
  -2.64/c/{Projections only}/memcolor/{Keep recipient KV}/ctxcolor/{Impose donor\\projections}/ctxcolor/context%
} {
  \node[anchor=west,text width=2.45cm,align=left,font=\footnotesize\bfseries]
    at (0,\y) {\heading};
  \node[action,draw=\base,fill=\base!8,text=\base] (\id-base)
    at (4.35,\y) {\operation};
  \node[action,draw=\coord,fill=\coord!12,text=\coord] (\id-coord)
    at (8.25,\y) {\coordtext};
  \node[report,draw=\out,fill=\out!10,text=\out] (\id-report)
    at (11.9,\y) {\texttt{\answer}};
  \draw[flow] (\id-base.east) -- (\id-coord.west);
  \draw[flow] (\id-coord.east) -- (\id-report.west);
}
\end{tikzpicture}
\caption{\textbf{Which part of the transplanted cache carries the report?}
Each row independently edits the same recipient's Q+A cache. Projections are components along the fitted directions. Schematic example; Table~\ref{tab:clamp_main} gives measured rates in both directions.}
\label{fig:clamp}
\end{figure}

Copying the Q+A cache makes the source report follow the donor (Section~\ref{sec:b2}). Steering also changes the report (Section~\ref{sec:b3}). Do the fitted directions capture the information that makes the transplant work? We pair true-fact and irrelevant-fact runs with identical Q+A tokens (Figure~\ref{fig:clamp}). We reuse the value directions from Section~\ref{sec:b3} and fit key directions from the same contrast. All edits cover Q+A at every layer, after answering and before asking for the source. The document and all dialogue tokens stay fixed.

\paragraph{Are these components needed for transfer?} Copy the donor's full Q+A cache, then restore its projections onto the fitted directions to those of the recipient. For a value vector, this \emph{clamp} is
\begin{equation}
 v_{\mathrm{clamp}} =
 \underbrace{v_D}_{\text{copy donor}}
 - \underbrace{\langle v_D,u\rangle u}_{\text{remove donor projection}}
 + \underbrace{\langle v_R,u\rangle u}_{\text{restore recipient projection}}.
 \label{eq:clamp_main}
\end{equation}
Here $v_D$ and $v_R$ are the original donor and recipient value vectors at the same layer and position; $u$ is the unit direction fitted for that layer and span. The projection $\langle v,u\rangle u$ is the component of $v$ along $u$, with $\langle\cdot,\cdot\rangle$ denoting the dot product. Keys use a norm-preserving rotation instead (Appendix~\ref{app:clamp}). Joint K+V clamps reduce flips from 100\% to \textbf{1.1\%} on OLMo and from 100\% to \textbf{0\%} on Qwen.

\paragraph{Is transferring these projections enough?} Start from the recipient's cache and impose the donor's projections: for values, exchange $D$ and $R$ on the right of Equation~\ref{eq:clamp_main}. Without amplification, this flips \textbf{97.9\% of OLMo and 100\% of Qwen reports}, nearly reproducing the full transplant.

\begin{table}[!htbp]
\caption{\textbf{Projections onto fitted directions block and reproduce report transfer.} Report-flip rates (\%) over directional evaluations for true-fact $\leftrightarrow$ irrelevant-fact pairs, pooled over two held-out relation splits. Every row tests the same pairs in both directions.}
\label{tab:clamp_main}
\centering\footnotesize\setlength{\tabcolsep}{5pt}
\begin{tabular}{@{}lrr@{}}
\toprule
Intervention & \olmoS{} & \qwenS{}\\
\midrule
Full Q+A KV transplant & 100\% & 100\%\\
\quad + restore recipient K+V projections & \textbf{1.1\%} & \textbf{0\%}\\
\quad + clamp energy-matched subspace & 100\% & 94.7\%\\
No transplant: donor K+V projections only & \textbf{97.9\%} & \textbf{100\%}\\
\bottomrule
\end{tabular}
\end{table}

\paragraph{Nearly necessary and sufficient for report transfer.} By contrast, clamping a control subspace with comparable donor--recipient difference energy on fitting data leaves 100\% and 94.7\% of reports flipped (Table~\ref{tab:clamp_main}). On these pairs, the fitted K+V components therefore largely account for the full transplant's effect on the report. Values-only tests and behavioural-check limits are in Appendix~\ref{app:clamp}.

\section{Steering values during answering changes the chosen source}
\label{sec:b4}

So far we edited the cache only after the answer was complete, which can change the report but not the answer. We now ask whether the same value-steering direction also affects which source the model answers from. To test this, we steer values while the model reads the question and generates its answer. This readout needs no source question. Given the document ``The capital of France is Tokyo'', the answer ``Tokyo'' comes from the document and ``Paris'' from memory.

\begin{examplebox}{Move the intervention before the answer is complete (schematic)}
\textbf{Document:} ``The capital of France is Tokyo.''\quad \textbf{Question:} capital of France?\\
\textbf{Edit during answering:} steer values as question and answer tokens are processed.\\
\textbf{Readout:} the answer switches between ``Tokyo'' and ``Paris''.
\end{examplebox}

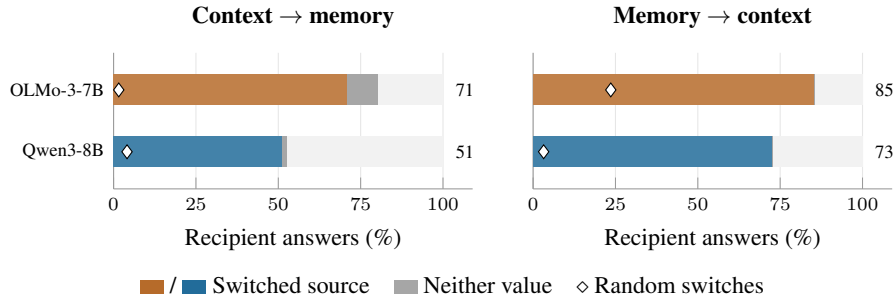
\begin{figure}[!htbp]
\centering
\begin{tikzpicture}
\begin{groupplot}[group style={group size=2 by 1,horizontal sep=0.8cm},
 paperaxis,scale only axis,width=4.75cm,height=1.8cm,
 xmin=0,xmax=109,xtick={0,25,50,75,100},xmajorgrids,
 ymin=-.6,ymax=1.6,ytick={0,1},
 yticklabels={Qwen3-8B,OLMo-3-7B},
 xlabel={Recipient answers (\%)}]
\nextgroupplot[title={Context $\to$ memory}]
\fill[black!5] (axis cs:0,0.75) rectangle (axis cs:100,1.25);
\fill[cOa!85] (axis cs:0,0.75) rectangle (axis cs:70.930233,1.25);
\fill[black!35] (axis cs:70.930233,0.75) rectangle (axis cs:80.232558,1.25);
\addplot[only marks,mark=diamond*,mark size=2.6pt,black,mark options={fill=white}] coordinates {(1.569767,1)};
\node[anchor=west,font=\scriptsize] at (axis cs:101,1) {71};
\fill[black!5] (axis cs:0,-0.25) rectangle (axis cs:100,0.25);
\fill[cQa!85] (axis cs:0,-0.25) rectangle (axis cs:51.219512,0.25);
\fill[black!35] (axis cs:51.219512,-0.25) rectangle (axis cs:52.439024,0.25);
\addplot[only marks,mark=diamond*,mark size=2.6pt,black,mark options={fill=white}] coordinates {(4.085366,0)};
\node[anchor=west,font=\scriptsize] at (axis cs:101,0) {51};
\nextgroupplot[title={Memory $\to$ context},yticklabels={,}]
\fill[black!5] (axis cs:0,0.75) rectangle (axis cs:100,1.25);
\fill[cOa!85] (axis cs:0,0.75) rectangle (axis cs:85.321101,1.25);
\fill[black!35] (axis cs:85.321101,0.75) rectangle (axis cs:85.321101,1.25);
\addplot[only marks,mark=diamond*,mark size=2.6pt,black,mark options={fill=white}] coordinates {(23.577982,1)};
\node[anchor=west,font=\scriptsize] at (axis cs:101,1) {85};
\fill[black!5] (axis cs:0,-0.25) rectangle (axis cs:100,0.25);
\fill[cQa!85] (axis cs:0,-0.25) rectangle (axis cs:72.789116,0.25);
\fill[black!35] (axis cs:72.789116,-0.25) rectangle (axis cs:72.789116,0.25);
\addplot[only marks,mark=diamond*,mark size=2.6pt,black,mark options={fill=white}] coordinates {(3.197279,0)};
\node[anchor=west,font=\scriptsize] at (axis cs:101,0) {73};
\end{groupplot}
\end{tikzpicture}
\par\smallskip\footnotesize
\textcolor{cOa}{\rule{9pt}{5pt}} / \textcolor{cQa}{\rule{9pt}{5pt}} Switched source\quad
\textcolor{black!35}{\rule{9pt}{5pt}} Neither value\quad
$\diamond$ Random switches
\caption{\textbf{The value intervention also changes the answer's source.} Value steering at $\alpha=2$. Left: instructed to use documents, but steering favours memory. Right: the instruction favours knowledge when the document may be wrong; steering favours the document. Coloured bars are switches; grey are answers containing neither value; diamonds are mean random switches over three draws. Switch percentages are on the right. Rates are over eligible item--split pairs. The residual-stream comparison is in Table~\ref{tab:b4}.}
\label{fig:source}
\end{figure}

We reuse the fitted value direction of Equation~\ref{eq:direction} (true-fact minus irrelevant-fact values), without refitting for the instructions tested here. We apply it at $\alpha=2$ at every layer, first to the question tokens and then to each generated answer token. To make the test stringent, the instruction always opposes the edit. We steer toward memory when the model is told to use the document, and toward the document when it is told the document may contain errors. Recipients are runs that initially answered from the source we steer away from.

Steering overrides the instruction in both directions on both models (Figure~\ref{fig:source}). Toward memory, it switches 71\% of OLMo and 51\% of Qwen answers; toward the provided context, 85\% and 73\%. In these two conditions, random directions of equal norm switch at most 4.1\% of answers, except for OLMo toward the context (24\%), where the fitted direction still switches 3.6 times as many. At most 9.3\% of answers are damaged. Residual-stream steering is weaker than value steering on OLMo and stronger on Qwen (Appendix~\ref{app:sourcefull}). The same fitted value direction therefore changes both the later source report and which source supplies the answer, even when the instruction opposes the intervention.

\section{Discussion and limitations}
\label{sec:discussion}

Together, the interventions give a consistent account of source reporting in this setting. Editing the document after answering leaves the report unchanged. Transplanting the question-and-answer cache moves the report while every recipient token stays fixed. A shared value direction in that cache can change the report. K+V projections onto fitted directions also nearly block or reproduce report transfer on true-fact--irrelevant-fact pairs. Applying the same value direction during answering changes which source supplies the answer. The report therefore depends on state computed before the source question is asked. These findings concern eligible runs and one source-question wording. Document edits and transplants extend to selected DynamicQA items; steering and projection tests use only 7--8B models. General attribution reliability in natural retrieval settings remains untested. Three questions remain open:

\paragraph{Meaning and formation of the directions.} Our interventions establish a causal role in reporting without resolving whether the trace represents evidence availability, actual source use, or both. The fixed-document comparison leaves this distinction open (Appendix~\ref{app:fixedcontext}). Future work should determine what these directions encode and how they form during computation. H1 and H2 fail as complete accounts, but hybrid mechanisms may contribute.

\paragraph{Roles of keys and values.} Joint K+V edits nearly block or reproduce report transfer. Value-only clamps are less effective when keys are also transplanted (Appendix~\ref{app:clamp}). Explaining how keys and values contribute separately, and how they interact, is a next step. Answer repetition and document read-back remain untested for these joint K+V edits.

\paragraph{Roles of question and answer states.} Our main interventions edit Q+A together. Separate-span ablations show that their contributions vary across models and conditions (Appendices~\ref{app:where} and~\ref{app:b3d}). Future work should distinguish what is established while processing the question from what is added during answering. The fitted directions span layers and Q/A positions; they do not define one global source variable.

\subsubsection*{Reproducibility statement}
The appendix reports the joint steering sweep, compact span and fitting-contrast comparisons, controls and the mapping from experiments to scripts and result artifacts. Full sweeps, random controls, behavioural checks and the fitting protocol are preserved in the accompanying data.

\clearpage
\subsubsection*{AI use statement}
We used generative AI tools to assist with drafting and editing the manuscript and with writing code. The authors take responsibility for the interpretation of the results, the scientific conclusions, and the final content of this paper.


\clearpage

\appendix

\section{Related work}
\label{app:related}

\paragraph{Reports about internal computation.} Prior work studies uncertainty, truth-related states and the faithfulness of explanations and self-reports \citep{kadavath2022language,lin2022teaching,azaria2023internal,turpin2023language,lanham2023measuring,binder2025looking}. In a close precedent, \citet{lindsey2025emergent} injects a concept before an artificial prefill, making Claude more likely to accept it as intentional. We study naturally generated answers and test the later source report with document edits and Q+A transplants that preserve the question and answer tokens.

\paragraph{Knowledge conflicts and source selection.} Contextual and parametric knowledge can compete \citep{longpre2021entity,xie2024adaptive,ortu2024competition,jin2024cutting}. \citet{zhao2024analysing} predict source choice from residual activations; SpARE \citep{zhao2025steering} controls it through sparse autoencoder features. Our central contribution concerns the mechanism of the subsequent source report. Section~\ref{sec:b4} connects source-related state to answer selection.

\paragraph{Attribution to context.} ContextCite \citep{cohenwang2024contextcite} estimates contextual contributions to generation using context ablations and a sparse linear surrogate. We instead intervene after answering, preserving downstream Q+A state. Invariance of the later report therefore does not imply that the document did not contribute to generation.

\paragraph{Patching and steering.} We build on activation patching \citep{vig2020investigating,meng2022locating,zhang2024towards}, mean-difference steering \citep{turner2023steering,zou2023representation,li2023inference,rimsky2024steering}, and KV-cache steering \citep{belitsky2025kv}, with matched controls to assess intervention specificity \citep{wu2025axbench}.

\clearpage
\section{Generated factual items}
\label{app:generated_items}

Each candidate is a subject--relation--object statement, rendered as ``The \{phrase\} of \{subject\} is \{value\}.'' and queried with ``\{Who/What\} is the \{phrase\} of \{subject\}?'' Table~\ref{tab:generated_items} describes the original pool of 155 candidates. Model- and intervention-specific eligibility checks determine which items enter each reported result (Section~\ref{sec:setup}).

\begin{table}[H]
\caption{\textbf{The eight relations in the candidate pool.} Counts precede the length-matching and eligibility requirements of the reported experiments.}
\label{tab:generated_items}
\centering\footnotesize\setlength{\tabcolsep}{12pt}
\begin{tabular}{@{}llr@{}}
\toprule
Relation & Phrase & Candidates\\
\midrule
capital & capital & 30\\
symbol & chemical symbol & 26\\
author & author & 22\\
continent & continent & 21\\
language & official language & 20\\
painter & painter & 12\\
composer & composer & 12\\
inventor & inventor & 12\\
\midrule
\textbf{Total} & & \textbf{155}\\
\bottomrule
\end{tabular}
\end{table}

\paragraph{Counterfactual construction.}
Counterfactuals use other attested values of the same relation. For candidate $i$, let \texttt{others} list that relation's distinct values excluding the true value. The counterfactual is chosen deterministically as $\texttt{others}[i\bmod|\texttt{others}|]$; the second counterfactual uses the next value in the same cycle. A capital is thus replaced by a capital and an author by an author, preserving the answer type without inventing new strings. Values need not be unique across subjects: the continent relation has six distinct values across 21 candidates.

\paragraph{Parametric-knowledge check.}
Each candidate is queried with neutral filler in place of an informative document. The check accepts an answer if it contains the true value and does not contain the counterfactual, using normalised string matching over at most 12 greedily decoded tokens. It is applied per model, so eligible item sets may differ. This tests recall of the factual answer; the additional answer and intact-report requirements for each intervention are specified in Section~\ref{sec:setup}.

\paragraph{Length matching.}
The seven document variants described in Appendix~\ref{app:documents}, including neutral fillers, have equal token counts within an item under each model's tokenizer. This keeps the later question and answer positions aligned across conditions, allowing cache entries to be exchanged without a positional shift. Length matching is an additional constraint on the candidate inventory, separate from the model's ability to recall a fact.

\paragraph{Relation-based splits.}
Directions are fitted on DEV relations and tested on held-out relations. Two candidate splits exchange the groups \{capital, continent, language, symbol\} and \{author, painter, composer, inventor\}; the third uses a mixed DEV group. The exact memberships and retained splits for the reported steering experiments are given in Appendix~\ref{app:strength}. Thus, evaluation relations are absent from the corresponding fitting set, while the extent of domain transfer depends on the split.

\clearpage
\section{Additional results and details}

\FloatBarrier
\Needspace{12\baselineskip}
\subsection{Documents and document interventions}
\label{app:documents}
The examples in Table~\ref{tab:docs} illustrate the conditions. Each item has seven documents: one true fact and two variants of each of the other conditions. Exact token-length equality is enforced within each item in the experimental dataset. Appendix~\ref{app:generated_items} describes the candidate facts and their construction.
\begin{table}[H]
\caption{The seven document variants for the question \emph{What is the capital of France?} Each control variant is a second document of the same condition.}
\label{tab:docs}
\centering\scriptsize\setlength{\tabcolsep}{4pt}
\begin{tabular}{@{}llll@{}}
\toprule
document & example & control variant & role\\
\midrule
true fact & The capital of France is Paris. & --- & relevant, true value\\
counterfactual & The capital of France is Tokyo. & \dots\ is Lima. & relevant, false value\\
irrelevant fact & The capital of Canada is Ottawa. & \dots\ of Egypt is Cairo. & true fact, other subject\\
neutral sentence & Snow melted slowly on the hills. & The library was open until late. & no fact at all\\
\bottomrule
\end{tabular}
\end{table}
\begin{table}[H]
\caption{Document edits leave the report fixed; question transplants often move it. Percentages of eligible items with the stated report. Bold: the weak direction on each 32B model (Section~\ref{sec:b2}).}
\label{tab:b1}
\centering\footnotesize\setlength{\tabcolsep}{2.5pt}
\begin{tabular}{@{}>{\raggedright\arraybackslash}p{2.5in}cccc@{}}
\toprule
 & \olmoS{} & \qwenS{} & \olmoL{} & \qwenL{}\\
\midrule
\multicolumn{5}{@{}l}{\emph{Counterfactual run (read ``Tokyo'', answered ``Tokyo''): reports \ctx{}}}\\
intact & 100\% & 100\% & 100\% & 100\%\\
\xDrop{} & 100\% & 100\% & 100\% & 100\%\\
\xDswap{} $\to$ irrelevant fact & 100\% & 100\% & 100\% & 100\%\\
\quad model quotes the new document & 100\% & 100\% & 100\% & 100\%\\
\midrule
\multicolumn{5}{@{}l}{\emph{Irrelevant-fact run (answered ``Paris'' from memory): reports \ctx{}}}\\
intact & 0\% & 0\% & 0\% & 0\%\\
\xDswap{} $\to$ true fact (its own answer) & 0\% & 0\% & 0\% & 0\%\\
\xDswap{} $\to$ counterfactual & 0\% & 0\% & 0\% & 0\%\\
\midrule
\multicolumn{5}{@{}l}{\emph{\xQswap{} (document and answer stay): report becomes \dots}}\\
counterfactual \from{} irrelevant-fact: \kno{} & 97.9\% & 100\% & 100\% & \textbf{14.2\%}\\
irrelevant-fact \from{} counterfactual: \ctx{} & 86.6\% & 90.6\% & \textbf{6.7\%} & 100\%\\
controls: report changes & 0\% & 0\% & 0\% & 0\%\\
\bottomrule
\end{tabular}
\end{table}
\paragraph{Read-back caveats.}
\label{app:caveats}
Qwen3, after receiving the question KV of an irrelevant-fact run, says the document does not mention the subject when asked to read it (77.4\% on \qwenS{}); on that model the clean dissociation is the counterfactual $\leftarrow$ neutral question-swap test (Table~\ref{tab:b2}). After the document's KV is dropped, \qwenL{} reconstructs the document from its own answer when asked to repeat it (26.2\%); the \xDswap{} is the clean version of that cell.

\FloatBarrier
\Needspace{26\baselineskip}
\subsection{Question and answer transplants}
\label{app:transplants}
\begin{figure}[!htbp]
\centering
\begin{tikzpicture}
\begin{groupplot}[group style={group size=2 by 1,horizontal sep=0.75cm},
 paperaxis,scale only axis,width=5.05cm,height=2.25cm,
 xmin=-3,xmax=113,xtick={0,25,50,75,100},xmajorgrids,
 ymin=-0.5,ymax=3.5,ytick={0,1,2,3},
 yticklabels={Qwen3-32B,OLMo-3.1-32B,Qwen3-8B,OLMo-3-7B},
 xlabel={Reports flipped (\%)}]
\nextgroupplot[title={Context $\to$ knowledge}]
\addplot[only marks,mark=*,mark size=2.7pt,color=cOa] coordinates {(97.938144,3)};
\node[anchor=west,font=\scriptsize] at (axis cs:99.938144,3) {98\%};
\addplot[only marks,mark=*,mark size=2.7pt,color=cQa] coordinates {(100.000000,2)};
\node[anchor=west,font=\scriptsize] at (axis cs:102.000000,2) {100\%};
\addplot[only marks,mark=*,mark size=2.7pt,color=cOb] coordinates {(100.000000,1)};
\node[anchor=west,font=\scriptsize] at (axis cs:102.000000,1) {100\%};
\addplot[only marks,mark=*,mark size=2.7pt,color=cQb] coordinates {(14.166667,0)};
\node[anchor=west,font=\scriptsize] at (axis cs:16.166667,0) {14\%};
\addplot[only marks,mark=x,mark size=3pt,thick,black!60] coordinates {(0,0) (0,1) (0,2) (0,3)};
\nextgroupplot[title={Knowledge $\to$ context},yticklabels={,,,}]
\addplot[only marks,mark=*,mark size=2.7pt,color=cOa] coordinates {(86.597938,3)};
\node[anchor=west,font=\scriptsize] at (axis cs:88.597938,3) {87\%};
\addplot[only marks,mark=*,mark size=2.7pt,color=cQa] coordinates {(90.566038,2)};
\node[anchor=west,font=\scriptsize] at (axis cs:92.566038,2) {91\%};
\addplot[only marks,mark=*,mark size=2.7pt,color=cOb] coordinates {(6.666667,1)};
\node[anchor=west,font=\scriptsize] at (axis cs:8.666667,1) {7\%};
\addplot[only marks,mark=*,mark size=2.7pt,color=cQb] coordinates {(100.000000,0)};
\node[anchor=west,font=\scriptsize] at (axis cs:102.000000,0) {100\%};
\addplot[only marks,mark=x,mark size=3pt,thick,black!60] coordinates {(0,0) (0,1) (0,2) (0,3)};
\end{groupplot}
\end{tikzpicture}
\par\smallskip\footnotesize $\times$ Document swap\qquad $\bullet$ Question KV swap
\caption{\textbf{The report moves with question state, not with the edited document.} Question swaps exchange counterfactual and irrelevant-fact runs; coloured dots show the fraction whose report follows the donor. Document swaps replace a counterfactual with an irrelevant fact (left), or an irrelevant fact with a true fact stating the answer (right): zero flips on every model. Document deletion also produces zero flips. Question swaps are asymmetric at 32B, motivating the answer-cache tests. Controls: Table~\ref{tab:b1}.}
\label{fig:localisation}
\end{figure}
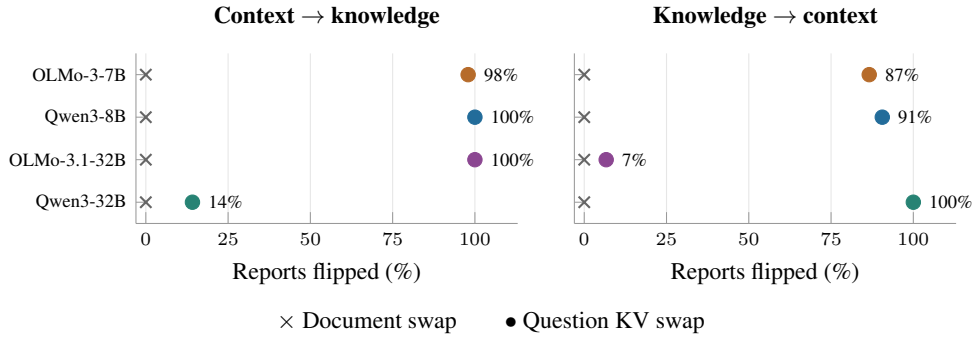

Question-only tests first located a signal downstream of the document, but were not uniformly effective. Figure~\ref{fig:localisation} retains that comparison. The main Q+A result uses identical-answer pairs, a different eligible subset from the counterfactual/memory pairs below.

Rates are over facts for which both runs qualify; the identical-text intervention additionally requires identical answer tokens. The read-back row measures the joint event: the report changes to \kno{} and the model can still read the false value from the document.

\begin{table}[H]
\caption{The report follows the cache even when the answer, or all visible text, is fixed. Percentages of eligible items whose report became the stated one. Bold: \qwenL{} with a false answer, where question-only transplants are weakest.}
\label{tab:b2}
\centering\footnotesize\setlength{\tabcolsep}{2.5pt}
\begin{tabular}{@{}>{\raggedright\arraybackslash}p{2.5in}cccc@{}}
\toprule
recipient \from{} donor: report becomes & \olmoS{} & \qwenS{} & \olmoL{} & \qwenL{}\\
\midrule
\multicolumn{5}{@{}l}{\emph{The same true answer ``Paris''; \xQswap{}}}\\
true-fact \from{} neutral: \kno{} & 80.2\% & 100\% & 100\% & 95.8\%\\
neutral \from{} true-fact: \ctx{} & 97.4\% & 100\% & 73.3\% & 100\%\\
\midrule
\multicolumn{5}{@{}l}{\emph{A false answer ``Tokyo'' in plain sight; \xQswap{}}}\\
counterfactual \from{} neutral: \kno{} & 90.4\% & 100\% & 68.3\% & \textbf{25.8\%}\\
\quad still reads ``Tokyo'' in the document & 83.3\% & 97.2\% & 68.3\% & 25.8\%\\
neutral \from{} counterfactual: \ctx{} & 89.5\% & 97.2\% & 49.2\% & 100\%\\
\midrule
\multicolumn{5}{@{}l}{\emph{\xIswap{}: question and answer swapped, visible text unchanged}}\\
true-fact \from{} neutral: \kno{} & 100\% & 100\% & 100\% & 100\%\\
neutral \from{} true-fact: \ctx{} & 100\% & 100\% & 100\% & 100\%\\
true-fact \from{} irrelevant-fact: \kno{} & 100\% & 100\% & 100\% & 100\%\\
irrelevant-fact \from{} true-fact: \ctx{} & 100\% & 100\% & 100\% & 100\%\\
controls: report changes & 0\% & 0\% & 0\% & 0\%\\
\bottomrule
\end{tabular}
\end{table}

\FloatBarrier
\Needspace{12\baselineskip}
\subsection{Keys and values in question-only transplants}
\label{app:kv}
\begin{table}[H]
\caption{The question span swapped as keys+values, keys only or values only. Cells: \olmoS{}\dt\qwenS{}, percentages of eligible items whose report flipped (remove or create the \ctx{} report). Value-only transplants create the report in most cases; removing it is more effective with keys and values together.}
\label{tab:b3a}
\centering\scriptsize\setlength{\tabcolsep}{3pt}
\begin{tabular}{@{}lccc@{}}
\toprule
swap of the question span & keys+values & keys only & values only\\
\midrule
counterfactual \from{} irrelevant (remove) & 97.9\%\dt 100\% & 0\%\dt 0\% & 6.2\%\dt 0\%\\
true-fact \from{} irrelevant (remove) & 92.9\%\dt 100\% & 0\%\dt 0\% & 0\%\dt 0\%\\
irrelevant \from{} counterfactual (create) & 86.6\%\dt 90.6\% & 40.2\%\dt 0.9\% & 99\%\dt 97.2\%\\
irrelevant \from{} true-fact (create) & 90.9\%\dt 98.3\% & 54.5\%\dt 5.2\% & 100\%\dt 100\%\\
\bottomrule
\end{tabular}
\end{table}

\FloatBarrier
\Needspace{12\baselineskip}
\subsection{Question and answer spans in transplants}
\label{app:where}
Using the pairs from the \xIswap{} experiment (Section~\ref{sec:b2}), we swap the question cache, the answer cache, or both. The joint result establishes sufficiency for transferring the report on these pairs; it does not establish that each span separately encodes an independent source variable.

\begin{table}[H]
\caption{Percentages of eligible items whose report moved, by swapped span.}
\centering\footnotesize\setlength{\tabcolsep}{4pt}
\begin{tabular}{@{}lccc@{}}
\toprule
 & question only & answer only & question $+$ answer\\
\midrule
\olmoS{}, \qwenS{}, \qwenL{} (all four directions) & 75--100\% & 0--3\% & 100\%\\
\olmoL{}, irrelevant-fact \from{} true-fact & 30\% & 0\% & 100\%\\
\olmoL{}, true-fact \from{} irrelevant-fact & 100\% & 59\% & 100\%\\
\bottomrule
\end{tabular}
\end{table}

\FloatBarrier
\Needspace{12\baselineskip}
\subsection{Joint report steering across strengths}
\label{app:strength}
The sweep uses the true-fact versus irrelevant-fact contrast of Section~\ref{sec:b3}. Rates are over eligible item--split pairs pooled over splits 0 and 2, not independent facts. The full fitted-direction sweep and random controls are retained in the accompanying data.

\paragraph{Relation splits.} The three candidate DEV sets are: split 0, \{capital, continent, language, symbol\}; split 1, \{author, painter, composer, inventor\}; and split 2, \{capital, language, author, composer\}. Evaluation uses the remaining relations. The reported Q+A sweep and the answer-only fitting-contrast comparison use splits 0 and 2. Exact prompts and fitting details are archived in \path{data/steering_protocol.json}.

\begin{table}[H]
\caption{\textbf{Joint Q+A steering across all four document conditions.} Report-flip rates (\%) over eligible item--split pairs pooled across the two retained relation splits. Every cell reaches 100\% at $\alpha=4$; at $\alpha=3$, the only exception is OLMo on true facts (98.9\%). Bold marks complete flipping. Random means are given below.}
\label{tab:b3b}
\centering\footnotesize\setlength{\tabcolsep}{5pt}
\begin{tabular}{@{}llrrrr@{}}
\toprule
Document / direction & Model & $\alpha=1$ & $\alpha=2$ & $\alpha=3$ & $\alpha=4$\\
\midrule
Irrelevant fact: create & \olmoS{} & 42.6\% & \textbf{100\%} & \textbf{100\%} & \textbf{100\%}\\
 & \qwenS{} & 4.9\% & \textbf{100\%} & \textbf{100\%} & \textbf{100\%}\\
\addlinespace[3pt]
Neutral sentence: create & \olmoS{} & 12.8\% & 88.5\% & \textbf{100\%} & \textbf{100\%}\\
 & \qwenS{} & 0\% & 2.4\% & \textbf{100\%} & \textbf{100\%}\\
\addlinespace[3pt]
Counterfactual: remove & \olmoS{} & 0\% & 60.5\% & \textbf{100\%} & \textbf{100\%}\\
 & \qwenS{} & 12.5\% & \textbf{100\%} & \textbf{100\%} & \textbf{100\%}\\
\addlinespace[3pt]
True fact: remove & \olmoS{} & 0\% & 46.6\% & 98.9\% & \textbf{100\%}\\
 & \qwenS{} & 0\% & 96.6\% & \textbf{100\%} & \textbf{100\%}\\
\bottomrule
\end{tabular}
\par\smallskip\raggedright\scriptsize
Random means: all Qwen cells are zero. For OLMo, irrelevant-fact Q+A gives 1.1\% at $\alpha=3$ and 1.6\% at $\alpha=4$; neutral Q+A gives 2.2\% at $\alpha=4$. All other joint cells are zero.
\end{table}

\FloatBarrier
\Needspace{12\baselineskip}
\subsection{Question and answer spans in report steering}
\label{app:b3d}
Table~\ref{tab:b3d} compares Q, A and Q+A at a common strength. The joint edit is more effective in every model--condition cell, supporting the choice of Q+A in the main text. It does not imply that either span has a single exclusive function.

\begin{table}[H]
\caption{\textbf{Joint editing is more effective than either span alone.} Report-flip rates (\%) over eligible item--split pairs at the same strength, $\alpha=3$, for all three edits. This descriptive comparison does not establish that Q and A encode independent source variables.}
\label{tab:b3d}
\centering\footnotesize\setlength{\tabcolsep}{6pt}
\begin{tabular}{@{}llrrr@{}}
\toprule
Document / direction & Model & Q only & A only & Q+A\\
\midrule
Irrelevant fact: create & \olmoS{} & 62.3\% & 67.2\% & \textbf{100\%}\\
 & \qwenS{} & 97.5\% & 3.7\% & \textbf{100\%}\\
\addlinespace[3pt]
Neutral sentence: create & \olmoS{} & 42.3\% & 62.8\% & \textbf{100\%}\\
 & \qwenS{} & 8.4\% & 0\% & \textbf{100\%}\\
\addlinespace[3pt]
Counterfactual: remove & \olmoS{} & 0\% & 72.1\% & \textbf{100\%}\\
 & \qwenS{} & 58\% & 85.2\% & \textbf{100\%}\\
\addlinespace[3pt]
True fact: remove & \olmoS{} & 0\% & 37.5\% & \textbf{98.9\%}\\
 & \qwenS{} & 40.9\% & 23.9\% & \textbf{100\%}\\
\bottomrule
\end{tabular}
\par\smallskip\raggedright\scriptsize
At $\alpha=3$, random means are zero except OLMo irrelevant-fact Q (0.5\%) and Q+A (1.1\%). Answer-only random controls are zero at all four strengths. Across the full sweep, the maximum random rate is 2.2\%.
\end{table}

The distinction persists at higher strength in the counterfactual setting: answer-only steering at $\alpha=4$ flips 86\% of OLMo and 90.9\% of Qwen reports, whereas Q+A flips 100\% on both models. Thus, the complete effect reported in the main text belongs to the joint intervention.

\FloatBarrier
\Needspace{12\baselineskip}
\subsection{Content checks after report steering}
\label{app:edits}
These checks use the same edited Q+A caches and eligible item--split pairs as the report-steering sweep (block 3i). The report, answer repetition, document quotation and counterfactual-value read-back are separate continuations; no evaluations are filtered after intervention. Table~\ref{tab:report_checks} shows the operating points used for the two main conditions.

\begin{table}[H]
\caption{\textbf{Report flips with successful answer repetition at the operating points.} Joint success rates (\%) over eligible item--split pairs for Q+A steering. Every evaluation flips and passes repetition; document quotation also succeeds throughout.}
\label{tab:report_checks}
\centering\footnotesize\setlength{\tabcolsep}{6pt}
\begin{tabular}{@{}lcrr@{}}
\toprule
Document / direction & $\alpha$ & \olmoS{} & \qwenS{}\\
\midrule
Irrelevant fact: create & 2 & 100\% & 100\%\\
Counterfactual: remove & 3 & 100\% & 100\%\\
\bottomrule
\end{tabular}
\end{table}

For answer repetition, ``Repeat the answer you just gave, exactly.'' succeeds if the response contains the original answer. For quotation, ``Repeat the Document above word for word.'' succeeds if the response contains the full document, ignoring case and a final period. Both checks succeed on every unedited baseline. After steering, quotation succeeds in every evaluation in Table~\ref{tab:report_checks}. For counterfactual documents, ``According to the Document above, $\langle q\rangle$ Reply with the exact word from the Document.'' also retrieves the false value in every case; success requires that value and no refusal such as ``not mentioned''. These checks establish preservation at the stated settings, not across every strength or intervention. The full supplied checks are archived in \path{data/report_steering_checks.json}.

\FloatBarrier
\Needspace{18\baselineskip}
\subsection{Supporting versus irrelevant and fixed-document fitting contrasts}
\label{app:fixedcontext}
We compare two fitting procedures while editing only answer values. The original contrast is supporting versus irrelevant documents. The fixed-document contrast changes the instruction: $P_D$ says to use a relevant document even against prior knowledge; $P_M$ warns that the document may be wrong and favours memory.

\paragraph{Construction.} Let CF and IRR denote counterfactual and irrelevant documents. At layer $\ell$ and span $s$, we fit
\begin{equation}
 d^{s,\mathrm{FD}}_\ell = \mathop{\mathrm{mean}}_{i\in\mathrm{DEV}}\!\left[
   \bigl(\overline V^s_{\ell,i}(\mathrm{CF},P_D)-\overline V^s_{\ell,i}(\mathrm{CF},P_M)\bigr)
   -\bigl(\overline V^s_{\ell,i}(\mathrm{IRR},P_D)-\overline V^s_{\ell,i}(\mathrm{IRR},P_M)\bigr)
 \right].
 \label{eq:fixed_document_direction}
\end{equation}
Here $\overline V$ averages values over span positions, retaining all KV heads. The second difference subtracts the prompt contrast measured with an irrelevant document. All four answers are generated greedily, up to eight tokens, and their caches are rebuilt from the prompt and generated answer. CF answers differ across prompts; IRR answers need not match either. Each answer is averaged over its own span, so answer content can enter this contrast. The original answer direction instead requires identical answer tokens.

\paragraph{Fitting and evaluation.} We retain items that use CF under $P_D$ and memory under $P_M$, and memory with IRR under both prompts, with matching source reports: 48.4\% of OLMo and 77.9\% of Qwen items. Both directions require at least ten DEV pairs. The answer-only comparison uses splits 0 and 2 (Appendix~\ref{app:strength}); split 1 fails the DEV-pair minimum for the fixed-document contrast on OLMo and the original contrast on Qwen. Recipients are eligible $P_D$ runs; edits cover answer values at every layer after generation. Exact prompts and selection rules are in \path{data/steering_protocol.json}.

\begin{table}[H]
\caption{\textbf{Two fitting contrasts, the same answer-only intervention.} Report-flip rates (\%) over eligible evaluations at $\alpha=3$; remove/create refer to \ctx{}. Supporting versus irrelevant is the original contrast; fixed document uses a prompt correction. This is a descriptive comparison at a common multiplier, not at matched edit norms or a separately optimised strength.}
\label{tab:fixedcontext}
\centering\footnotesize\setlength{\tabcolsep}{6pt}
\begin{tabular}{@{}llrr@{}}
\toprule
Document / direction & Model & \shortstack[r]{Supporting vs.\\irrelevant} & \shortstack[r]{Fixed\\document}\\
\midrule
True fact: remove & \olmoS{} & 37.5\% & 30.7\%\\
 & \qwenS{} & 23.9\% & 77.3\%\\
\addlinespace[3pt]
Counterfactual: remove & \olmoS{} & 72.1\% & 72.1\%\\
 & \qwenS{} & 85.2\% & 100\%\\
\addlinespace[3pt]
Irrelevant fact: create & \olmoS{} & 67.2\% & 37.7\%\\
 & \qwenS{} & 3.7\% & 2.5\%\\
\addlinespace[3pt]
Neutral sentence: create & \olmoS{} & 62.8\% & 23.1\%\\
 & \qwenS{} & 0\% & 0\%\\
\bottomrule
\end{tabular}
\par\smallskip\raggedright\scriptsize
All norm-matched random controls are zero at every tested strength ($\alpha=1,2,3,4$), separately for each contrast. The full fitted-direction sweep is retained in the accompanying data.
\end{table}

The fixed-document direction removes \ctx{} more effectively on Qwen. On OLMo, neither contrast dominates removal across the full sweep, while the original contrast creates \ctx{} more effectively. Neither answer-only direction reliably creates \ctx{} on Qwen. These names describe fitting conditions, not established semantic variables.

\paragraph{Geometry and scope.} Norms are $\sqrt{\sum_\ell\|d_\ell\|^2}$ over all layers and heads, computed per split. Cosines are unweighted means of per-layer cosines, including a near-zero layer-0 contribution: 0.565 and 0.571 for OLMo, and 0.577 and 0.569 for Qwen, on splits 0 and 2, respectively. Per-split norms and full sweeps are archived with the protocol and comparison data. Answer repetition was not measured at $\alpha=3$, so Table~\ref{tab:fixedcontext} establishes report control at that strength, not preservation of answer readout.

\FloatBarrier
\Needspace{26\baselineskip}
\subsection{Source-choice steering: instructions and controls}
\label{app:sourcefull}
We use the two settings in Section~\ref{sec:b4}. To steer toward memory, we keep the document-priority instruction used in the preceding experiments. To steer toward the document, we instead warn that the document may contain errors and instruct the model to prefer its own knowledge. This provides recipients that initially answer from memory; otherwise, the model usually already uses the document. In both settings, recipients initially answer from the source we steer away from.

Value steering reuses the fitted directions of Section~\ref{sec:b3}, at all layers and without refitting for the instruction. The residual-stream comparison adds a mean hidden-state difference fitted at question positions under a neutral instruction. It uses one layer selected on fitting relations: OLMo layers 12, 16 and 12 across the three splits, Qwen layer 20, out of 32 and 36 layers respectively. Both methods intervene during question processing and answer generation at $\alpha=2$.

\begin{table}[H]
\caption{\xSsteer{} in the two main settings at $\alpha=2$. All rates are percentages over eligible item--split pairs. \emph{Switched}: the answer changes to the target source's value. \emph{Random}: source switches under random directions, averaged over three draws. \emph{Damaged}: the answer contains neither the document nor the memory value.}
\label{tab:b4}
\centering\footnotesize\setlength{\tabcolsep}{2.5pt}
\begin{tabular}{@{}llrrrrrr@{}}
\toprule
 & & \multicolumn{3}{c}{\olmoS{}} & \multicolumn{3}{c}{\qwenS{}}\\
\cmidrule(lr){3-5}\cmidrule(l){6-8}
Steer toward & Edit & Switched & Random & Damaged & Switched & Random & Damaged\\
\midrule
Memory & Values & 70.9 & 1.6 & 9.3 & 51.2 & 4.1 & 1.2\\
 & Residual & 16.3 & 0.2 & 0.6 & 75.6 & 3.7 & 2.4\\
\addlinespace[3pt]
Document & Values & 85.3 & 23.6 & 0 & 72.8 & 3.2 & 0\\
 & Residual & 52.3 & 5.5 & 1.8 & 98 & 1.2 & 0\\
\bottomrule
\end{tabular}
\end{table}

\FloatBarrier
\Needspace{12\baselineskip}
\subsection{Terminology used in the tables}
\emph{Flip}: the report changed to the other source label. \emph{Switched} (Section~\ref{sec:b4}): the answer changed to the other source's value. \emph{Damaged}: the answer contains neither value. \emph{Random}: a random direction with the same norm per layer, same positions and $\alpha$, averaged over 3 draws. \emph{Held-out}: directions fitted on some relations and measured on others; retained splits depend on the experiment.

\FloatBarrier
\Needspace{12\baselineskip}
\subsection{Where every number comes from}
\label{app:repro}
\begin{table}[H]
\caption{Scripts and result artifacts behind each section (per model, under a directory named by the model tag).}
\centering\scriptsize\setlength{\tabcolsep}{3pt}
\begin{tabular}{@{}>{\raggedright\arraybackslash}p{0.95in}>{\raggedright\arraybackslash}p{1.95in}>{\raggedright\arraybackslash}p{2.4in}@{}}
\toprule
section & script & artifacts\\
\midrule
items & \texttt{90\_clean\_items.py} & \texttt{items.jsonl}\\
\ref{sec:b1}--\ref{sec:b2}, document and question swaps & \texttt{91\_clean\_kv\_patching.py} & \texttt{kv\_patching.json}, \texttt{kv\_patching\_items.jsonl}\\
\ref{sec:b2}, \xIswap{} & \texttt{93\_clean\_answer\_patching.py} & \texttt{answer\_patching.json}, \texttt{answer\_patching\_items.jsonl}\\
tables for \ref{sec:b1}--\ref{sec:b2} & \texttt{92\_clean\_tables.py} & \texttt{tables.md}\\
\ref{sec:b3}, report-steering sweep & author-supplied aggregate table & \path{data/report_steering_update.csv}\\
Appendix~\ref{app:edits} & author-supplied block 3i (script 105) & \path{data/report_steering_checks.json}\\
\ref{sec:b4} & \texttt{96\_clean\_source\_steering.py} & \texttt{source\_steering.json}, \texttt{source\_steering\_items.jsonl}\\
Appendix~\ref{app:dynamicqa} & author-supplied aggregate tables & \path{data/dynamicqa_results.json}\\
Appendix~\ref{app:fixedcontext} & author-supplied blocks 3e/3g (scripts 100 and 102) & \path{data/answer_direction_comparison.json}; \path{data/steering_protocol.json}\\
\ref{sec:clamp}, Appendix~\ref{app:clamp} & author-supplied blocks 3j--3l & \path{data/transplant_clamp_results.json}\\
\bottomrule
\end{tabular}
\end{table}

\clearpage
\section{Clamps and transfers along fitted directions}
\label{app:clamp}

\subsection{Paired runs, directions and readout}
Blocks 3j (values), 3k (joint K+V) and 3l (energy controls) use \olmoS{}-Instruct and \qwenS{} with thinking disabled, bf16 and Transformers 4.57.1. OLMo and Qwen have 32 and 36 layers, respectively, with 32 and 8 KV heads of dimension 128.

Pairs follow Section~\ref{sec:b2}: a true-fact run reports \ctx{}, while its irrelevant-document partner gives identical answer tokens but reports \kno{}. Equal document lengths align positions. The question span runs from the document's end to the prompt's end, including chat-template tokens; the answer span ends at the answer's last token. Edits cover Q+A at every layer after answering and before the fixed source question. Other cache entries remain unchanged.

Directions are fitted on DEV relations and evaluated on held-out relations in two splits. Rates pool both splits and both directions for each true-fact--irrelevant-fact pair; the unit is a \emph{directional evaluation}, not a distinct fact. The readout is forced-choice $P(\ctx{})$; a flip crosses 0.5 away from the recipient's original report. These results are not pooled with the report-steering sweep.

For each layer and span, take the mean over DEV items of the position-averaged true-fact minus irrelevant-fact difference, then normalise the resulting direction. Value directions are the same as in Section~\ref{sec:b3}, flattened across KV heads: one coordinate per position in 4096 dimensions on OLMo and 1024 on Qwen. Key directions are fitted in the pre-RoPE, post-\texttt{k\_norm}-gain space. Grouping follows the model's key normalisation: one 4096-dimensional group on OLMo, but eight 128-dimensional head groups on Qwen. Question and answer positions use their respective fitted directions. Layer 0 has zero difference and is left unchanged. This is a family of local coordinates, not a rank-one intervention on the entire cache.

\subsection{Two operations with the same coordinates}
Let $u$ be a fitted unit direction, $x$ the vector being edited, and $s$ the paired vector supplying the target scalar projection (called a \emph{coordinate} here). A value edit is
\begin{equation}
  x' = x + \bigl(\langle s,u\rangle-\langle x,u\rangle\bigr)u .
  \label{eq:value_clamp}
\end{equation}
For a \emph{clamp}, first transplant Q+A K+V, then use the recipient's original vector as $s$. For \emph{donor coordinates only}, start from the recipient's cache and use the donor as $s$. The value component orthogonal to $u$ is unchanged. Both operations use observed coordinates rather than an amplified mean-difference shift.

\paragraph{Keys require a norm-preserving edit.}
We first undo RoPE using the relevant layer's rotary cosine and sine, including OLMo's distinct sliding-window and full-attention rotary modules:
\begin{equation}
 y=\frac{k\odot C-\operatorname{rot\_half}(k)\odot S}{C^2+S^2}.
 \label{eq:undo_rope_clamp}
\end{equation}
Reported round-trip error is approximately $2\times10^{-7}$. Within each key group, write $c=\langle y,u\rangle$, $r=y-cu$, and set $c'$ to the source coordinate, clipped to $\pm0.999\|y\|$. The edit is
\begin{equation}
 y'=c'u+r\,\frac{\sqrt{\|y\|^2-(c')^2}}{\|r\|}.
 \label{eq:key_clamp}
\end{equation}
We then reapply RoPE and store the result in bf16. This keeps the starting key norm while changing its orientation; unlike the value edit, it rescales the orthogonal component. It does not preserve attention weights. Earlier additive key interventions caused nonspecific report changes even for random directions, motivating the norm constraint.

The rotation is performed \emph{after} the key-normalisation gain. OLMo has near-zero gains in approximately a quarter of key dimensions, making inversion of the gain numerically unstable. The edit preserves Euclidean norm in this post-gain space, but need not stay on the exact ellipsoid defined by the model's normalisation and gain.

\clearpage
\subsection{Value and key contributions}
\begin{table}[H]
\caption{\textbf{Value-coordinate interventions.} Author block 3j. Report-flip rates (\%) over directional evaluations on true-fact--irrelevant-fact pairs. Create/remove refer to \ctx{}. Random-control rates are means over three draws.}
\label{tab:clamp_3j}
\centering\scriptsize\setlength{\tabcolsep}{5pt}
\begin{tabular}{@{}p{2.35in}rrrr@{}}
\toprule
& \multicolumn{2}{c}{\olmoS{}} & \multicolumn{2}{c}{\qwenS{}}\\
\cmidrule(lr){2-3}\cmidrule(l){4-5}
& Create & Remove & Create & Remove\\
\midrule
K+V transplant & 100\% & 100\% & 100\% & 100\%\\
V transplant & 100\% & 10.6\% & 100\% & 47.4\%\\
V transplant + V clamp & 0\% & 0\% & 0\% & 0\%\\
V transplant + random V clamp & 100\% & 10.6\% & 100\% & 47.4\%\\
K transplant & 78.7\% & 0\% & 39.5\% & 0\%\\
K+V transplant + V clamp & 100\% & 0\% & 68.4\% & 0\%\\
Donor V coordinate only & 100\% & 2.1\% & 100\% & 31.6\%\\
\bottomrule
\end{tabular}
\end{table}

Value-coordinate clamps eliminate flips produced by value-only transplants: 0\% of reports flip on either model, compared with 55.3\% on OLMo and 73.7\% on Qwen without the clamp. With donor keys also transplanted, a value-only clamp is less effective, leaving 50\% and 34.2\% of reports flipped, all when creating \ctx{}.

\begin{table}[H]
\caption{\textbf{Joint key/value interventions.} Author block 3k. Report-flip rates (\%) over directional evaluations on true-fact--irrelevant-fact pairs. Create/remove refer to \ctx{}. Random-control rates are means over three draws.}
\label{tab:clamp_3k}
\centering\scriptsize\setlength{\tabcolsep}{5pt}
\begin{tabular}{@{}p{2.35in}rrrr@{}}
\toprule
& \multicolumn{2}{c}{\olmoS{}} & \multicolumn{2}{c}{\qwenS{}}\\
\cmidrule(lr){2-3}\cmidrule(l){4-5}
& Create & Remove & Create & Remove\\
\midrule
K+V transplant + V clamp & 100\% & 0\% & 68.4\% & 0\%\\
K+V transplant + K+V clamp & 2.1\% & 0\% & 0\% & 0\%\\
K+V transplant + V clamp + random K clamp & 100\% & 0\% & 67.6\% & 0\%\\
K transplant + K clamp & 0\% & 0\% & 0\% & 0\%\\
K transplant + random K clamp & 76\% & 0\% & 38.7\% & 0\%\\
Random donor coordinates only & 0\% & 0\% & 0\% & 0\%\\
\bottomrule
\end{tabular}
\end{table}

Joint K+V clamps leave 1.1\% of reports flipped on OLMo and 0\% on Qwen. Residual OLMo flips occur only in the creation direction (2.1\%); no flips remain in the removal direction on either model.

Transferring only the donor's K+V coordinates flips 97.9\% of OLMo and 100\% of Qwen reports, pooled over both directions (Table~\ref{tab:clamp_controls}), matching the main result in Table~\ref{tab:clamp_main}.

\clearpage
\subsection{Energy-matched controls}
A random one-dimensional subspace captures very little of the donor--recipient difference. To test whether the selected coordinates matter simply because they capture a large change, block 3l also fits a control subspace on DEV. After projecting out the selected direction, it takes the fewest leading principal directions of per-position DEV differences needed to capture at least as much squared difference energy as the selected direction. This choice is made separately for each layer, span and key group. Reported median subspace dimensions range from 3--72 on OLMo and 6--186 on Qwen, versus one selected direction per group. The top-PC control uses only the first such principal direction. Random controls use three draws and the same intervention locations and coordinate-replacement geometry.

\begin{table}[H]
\caption{\textbf{Controls for coordinate transfer.} Author block 3l; true-fact--irrelevant-fact pairs only. All cells give report-flip rates (\%); random conditions average three draws. Top PC and matched subspaces are fitted after removing the selected direction.}
\label{tab:clamp_controls}
\centering\footnotesize\setlength{\tabcolsep}{5pt}
\begin{tabular}{@{}lrr@{}}
\toprule
Intervention & \olmoS{} & \qwenS{}\\
\midrule
Full K+V transplant & 100\% & 100\%\\
\quad + selected-coordinate clamp & 1.1\% & 0\%\\
\quad + top-PC clamp & 100\% & 100\%\\
\quad + energy-matched clamp & 100\% & 94.7\%\\
\quad + random clamp & 100\% & 100\%\\
\midrule
V transplant + selected-coordinate clamp & 0\% & 0\%\\
\quad + energy-matched clamp instead & 51.1\% & 65.8\%\\
\quad + random clamp instead & 55.3\% & 74.1\%\\
\midrule
Donor K+V coordinates only: selected & 97.9\% & 100\%\\
\quad top PC instead & 0\% & 0\%\\
\quad energy-matched subspace instead & 0\% & 2.6\%\\
\quad random directions instead & 0\% & 0\%\\
\bottomrule
\end{tabular}
\end{table}

\begin{table}[H]
\caption{\textbf{Held-out difference energy covered by each intervention subspace.} Fractions of squared donor--recipient differences, summed over Q+A positions and layers. Matching is on fitting data, so held-out fractions need not be equal.}
\label{tab:clamp_energy}
\centering\footnotesize\setlength{\tabcolsep}{5pt}
\begin{tabular}{@{}llrrr@{}}
\toprule
Model & Cache & Selected & Matched & Random\\
\midrule
\olmoS{} & V & 11.6\% & 12.3\% & 0.0\%\\
\olmoS{} & K & 25.1\% & 21.0\% & 0.0\%\\
\qwenS{} & V & 15.7\% & 18.5\% & 0.1\%\\
\qwenS{} & K & 36.2\% & 49.3\% & 0.8\%\\
\bottomrule
\end{tabular}
\end{table}

For a subspace projector $P$, the reported held-out energy fraction is
\begin{equation}
 \frac{\sum_{\ell,t}\|P_{\ell,s(t)}
       (x^{\mathrm{donor}}_{\ell,t}-x^{\mathrm{recipient}}_{\ell,t})\|^2}
      {\sum_{\ell,t}\|x^{\mathrm{donor}}_{\ell,t}
       -x^{\mathrm{recipient}}_{\ell,t}\|^2},
 \label{eq:clamp_energy}
\end{equation}
with key-group contributions also summed. This measures projected difference energy, not the norm of the nonlinear key edit. Matching is enforced on DEV only: the held-out OLMo key control captures 21.0\%, compared with 25.1\% for the selected directions. Qwen's matched control captures more held-out energy in both values and keys, yet leaves 94.7\% of reports flipped after clamping and flips only 2.6\% when its donor coordinates are imposed alone.

\paragraph{Scope of the behavioural checks.}
Answer repetition and document read-back were reported as fully preserved for \emph{value-only donor-coordinate} edits, but have not been supplied for the joint K+V conditions. Fixed dialogue tokens therefore do not establish preservation of subsequent content readouts. Nor do these interventions establish exclusive coordinate semantics or a complete account of ordinary source reporting.

\clearpage
\section{Replication on DynamicQA Wikipedia paragraphs}
\label{app:dynamicqa}

We extend the document and cache-transplant experiments to a Wikipedia-paragraph adaptation of DynamicQA \citep{marjanovic2024dynamicqa}. Each item has seven length-matched document variants. For each model, we restrict the item pool to questions it can answer from memory. This evaluates the interventions on richer, benchmark-derived passages while preserving a known memory answer.

Rates are over eligible items for each condition or pair, rather than item--split pairs or the full memory-known pools. ``Uninformative'' denotes the condition providing no answer information; ``irrelevant'' denotes a paragraph about a different fact. Identical-token transplants additionally require matching question and answer tokens. All four model columns use the same order as the main tables.

\subsection{Document interventions}
\label{app:dynamicqa_documents}
We edit the document after answering and retain the Q+A cache. Table~\ref{tab:dynamicqa_documents} reports the percentage of \ctx{} reports, rather than using a single flip metric for interventions with different initial reports.

\begin{table}[H]
\caption{\textbf{DynamicQA: document edits mostly preserve the source report.} Cells give \ctx{} report rates (\%) among eligible items, with Q+A retained. In the first two rows, \ctx{} is the original report; in the last two, the original report is \kno{}.}
\label{tab:dynamicqa_documents}
\centering\footnotesize\setlength{\tabcolsep}{3pt}
\begin{tabular}{@{}>{\raggedright\arraybackslash}p{2.35in}cccc@{}}
\toprule
Document edit & \olmoS{} & \qwenS{} & \olmoL{} & \qwenL{}\\
\midrule
Drop counterfactual & 100\% & 98.4\% & 99.4\% & 100\%\\
Counterfactual $\to$ irrelevant & 100\% & 100\% & 99.4\% & 100\%\\
Irrelevant $\to$ true fact & 0\% & 0\% & 0\% & 0\%\\
Irrelevant $\to$ counterfactual & 0\% & 0\% & 0\% & 0\%\\
\bottomrule
\end{tabular}
\end{table}

After deletion or irrelevant replacement of a counterfactual paragraph, \ctx{} is retained in 98.4--100\% of items across model--edit cells. Inserting a true or counterfactual paragraph into an irrelevant-document run creates no \ctx{} reports. Thus, the pattern from Section~\ref{sec:b1} extends to these Wikipedia paragraphs, with rare exceptions to exact invariance.

\subsection{Question-only transplants}
\label{app:dynamicqa_question}
Table~\ref{tab:dynamicqa_question} combines the question-only tests with irrelevant and uninformative donors. These effects vary by model and direction. The joint read-back rate uses all eligible items; it is not a success rate conditional on the report having flipped.

\begin{table}[H]
\caption{\textbf{DynamicQA: question-only transplants.} Cells give the stated report rate (\%) over eligible items; the recipient keeps its document and answer cache. The read-back row measures the joint event: the report becomes \kno{} and the model still reads the counterfactual value from its document.}
\label{tab:dynamicqa_question}
\centering\footnotesize\setlength{\tabcolsep}{3pt}
\begin{tabular}{@{}>{\raggedright\arraybackslash}p{2.55in}cccc@{}}
\toprule
Recipient $\leftarrow$ Q donor: report & \olmoS{} & \qwenS{} & \olmoL{} & \qwenL{}\\
\midrule
Counterfactual $\leftarrow$ irrelevant: \kno{} & 54.5\% & 100\% & 95.5\% & 57.6\%\\
Irrelevant $\leftarrow$ counterfactual: \ctx{} & 78.8\% & 72.4\% & 15.2\% & 72.9\%\\
\addlinespace[3pt]
True fact $\leftarrow$ uninformative: \kno{} & 94.9\% & 100\% & 99.5\% & 70.8\%\\
Uninformative $\leftarrow$ true fact: \ctx{} & 84.7\% & 77.1\% & 25.9\% & 59.8\%\\
\addlinespace[3pt]
Counterfactual $\leftarrow$ uninformative: \kno{} & 79.7\% & 99.2\% & 94.5\% & 16.8\%\\
\quad + reads the false document value & 66.7\% & 90.7\% & 93.9\% & 11.9\%\\
Uninformative $\leftarrow$ counterfactual: \ctx{} & 84.1\% & 72.9\% & 12.3\% & 61.6\%\\
\bottomrule
\end{tabular}
\end{table}

\clearpage
\subsection{Question and answer contributions with identical tokens}
\label{app:dynamicqa_spans}
For these pairs, the recipient's document and all question and answer tokens stay fixed. Only the cached state of Q, A, or Q+A is replaced. Each three-row group uses the same item subset, allowing a direct comparison of the edited spans. These subsets differ from the broader question-only tests above.

\begin{table}[H]
\caption{\textbf{DynamicQA: joint Q+A transplants transfer nearly every report with identical text.} Cells give donor-following report rates (\%) among eligible items. Within each pair, Q, A and Q+A edits use the same subset with identical question and answer tokens. Bold highlights the joint intervention, including its exceptions.}
\label{tab:dynamicqa_spans}
\centering\footnotesize\setlength{\tabcolsep}{3pt}
\begin{tabular}{@{}lcccc@{}}
\toprule
Edited span & \olmoS{} & \qwenS{} & \olmoL{} & \qwenL{}\\
\midrule
\multicolumn{5}{@{}l}{\emph{True fact $\leftarrow$ uninformative: report becomes \kno{}}}\\
Q only & 97.6\% & 100\% & 99.4\% & 77.6\%\\
A only & 10.6\% & 8.5\% & 51.9\% & 7.2\%\\
Q+A & \textbf{100\%} & \textbf{100\%} & \textbf{100\%} & \textbf{100\%}\\
\midrule
\multicolumn{5}{@{}l}{\emph{Uninformative $\leftarrow$ true fact: report becomes \ctx{}}}\\
Q only & 82.4\% & 84.5\% & 27.1\% & 72.4\%\\
A only & 1.2\% & 0\% & 0\% & 3.9\%\\
Q+A & \textbf{100\%} & \textbf{97.2\%} & \textbf{99.4\%} & \textbf{100\%}\\
\midrule
\multicolumn{5}{@{}l}{\emph{True fact $\leftarrow$ irrelevant: report becomes \kno{}}}\\
Q only & 60.9\% & 100\% & 100\% & 97.3\%\\
A only & 6.5\% & 19.2\% & 60.6\% & 19.8\%\\
Q+A & \textbf{100\%} & \textbf{100\%} & \textbf{100\%} & \textbf{100\%}\\
\midrule
\multicolumn{5}{@{}l}{\emph{Irrelevant $\leftarrow$ true fact: report becomes \ctx{}}}\\
Q only & 87\% & 79.8\% & 32.4\% & 80.2\%\\
A only & 28.3\% & 0\% & 0\% & 2.7\%\\
Q+A & \textbf{97.8\%} & \textbf{99\%} & \textbf{100\%} & \textbf{100\%}\\
\bottomrule
\end{tabular}
\end{table}

Joint Q+A transplants make 97.2--100\% of reports follow the donor across the 16 model--direction cells. The effect is nearly complete, rather than universal as on the templated task. The separate-span results again depend on model and direction, supporting the main paper's use of the joint span. This replication tests document edits and transplants; steering on DynamicQA remains outside the reported experiments.

\end{document}